\documentclass[sn-mathphys,Numbered]{sn-jnl}

\usepackage{makecell}
\usepackage{graphicx}%
\usepackage{multirow}%
\usepackage{amsmath,amssymb,amsfonts}%
\usepackage{amsthm}%
\usepackage{mathrsfs}%
\usepackage[title]{appendix}%
\usepackage{xcolor}%
\usepackage{textcomp}%
\usepackage{manyfoot}%
\usepackage{booktabs}%
\usepackage{algorithm}%
\usepackage{algorithmicx}%
\usepackage{algpseudocode}%
\usepackage{listings}%
\usepackage{geometry} 

\theoremstyle{thmstyleone}%
\theoremstyle{thmstyletwo}%

\theoremstyle{thmstylethree}%

\begin{document}

\title[Article Title]{Efficiently improving key weather variables forecasting by performing the guided iterative prediction in latent space}


\author[1,2]{\fnm{Shuangliang} \sur{Li}}\email{whu\_lsl@whu.edu.cn}

\author*[1,2]{\fnm{Siwei} \sur{Li}}\email{siwei.li@whu.edu.cn}


\affil*[1]{\orgdiv{Hubei Key Laboratory of Quantitative Remote Sensing of Land and Atmosphere}, \orgname{School of Remote Sensing and Information Engineering, Wuhan University}, \city{Wuhan}, \postcode{430079}, \country{China}}

\affil[2]{\orgdiv{Hubei Luojia Laboratory}, \orgname{Wuhan University}, \city{Wuhan}, \postcode{430079}, \country{China}}



\abstract{Weather forecasting refers to learning evolutionary patterns of some key upper-air and surface variables which is of great significance. Recently, deep learning-based methods have been increasingly applied in the field of weather forecasting due to their powerful feature learning capabilities. However, prediction methods based on the original space iteration struggle to effectively and efficiently utilize large number of weather variables. 
Therefore, we propose an 'encoding-prediction-decoding' prediction network. This network can efficiently benefit to more related input variables with key variables, that is, it can adaptively extract key variable-related low-dimensional latent feature from much more input atmospheric variables for iterative prediction. And we construct a loss function to guide the iteration of latent feature by utilizing multiple atmospheric variables in corresponding lead times. The obtained latent features through iterative prediction are then decoded to obtain the predicted values of key variables in multiple lead times. In addition, we improve the HTA algorithm in \cite{bi2023accurate} by inputting more time steps to enhance the temporal correlation between the prediction results and input variables. Both qualitative and quantitative prediction results on ERA5 dataset validate the superiority of our method over other methods. (The code will be available at \href{https://github.com/rs-lsl/Kvp-lsi}{https://github.com/rs-lsl/Kvp-lsi})}

\keywords{key variables prediction, latent space iterative prediction, HTAMI}



\maketitle

\section{Introduction}\label{sec1}

\begin{figure*}[t]
\centering
    \includegraphics[width=0.9\textwidth,trim=60 120 110 80,clip]{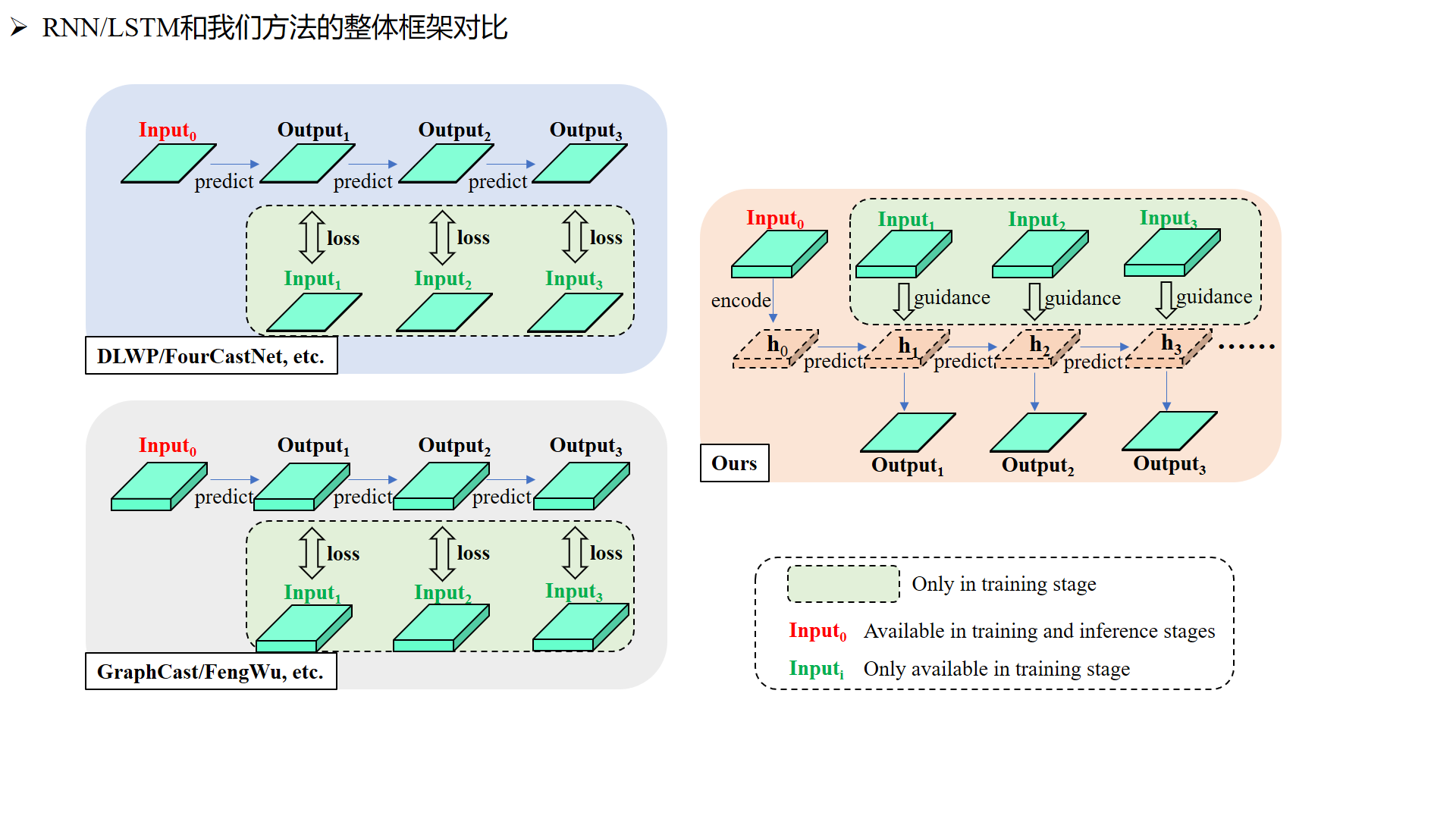}
    \caption{The comparison between the general DLWP \cite{https://doi.org/10.1029/2020MS002109}/FourCastNet \cite{pathak2022fourcastnet}, GraphCast \cite{doi:10.1126/science.adi2336}/FengWu \cite{chen2023fengwupushingskillfulglobal} and the proposed network framework. During the training phase, atmospheric variables with multiple lead times can be utilized to optimize the network parameters, such as the \textcolor{green}{$\rm{Input_1}$}, \textcolor{green}{$\rm{Input_2}$}, etc. Notably, our method allows for the input of more atmospheric variables to improve the prediction accuracy of key variables. Therefore, the input block in our framework is thicker than the output, while for the framework in left part, the input must has the same thickness as output.
}
    \label{fig:overall_frame}
\end{figure*}

Weather forecasting refers to predicting atmospheric and surface variables in multiple lead times, including some key variables such as 2m temperature, mean sea level pressure, geopotential and temperature at some specific pressure levels. These variables are pretty crucial for disaster prevention and resource management \cite{hua2024weather}\cite{https://doi.org/10.1029/2019MS001705}\cite{https://doi.org/10.1029/2020MS002109}\cite{https://doi.org/10.1029/2022MS003211}\cite{https://doi.org/10.1029/2020MS002405}. Therefore, accurately predicting these variables is of great significance.

Traditional numerical weather prediction (NWP) models through solving a series of partial differential equations to calculate changes in atmospheric variables. However, NWP requires complex physical equations to predict atmospheric variables. Additionally, it needs to run on supercomputers for several hours to obtain a forecast result, which consumes a lot of computational resources and time \cite{bi2023accurate}\cite{chen2023fuxi}\cite{doi:10.1126/science.adi2336}\cite{pathak2022fourcastnet}.

\begin{figure*}[t]
\centering
    \includegraphics[width=1\textwidth,trim=0 80 0 0,clip]{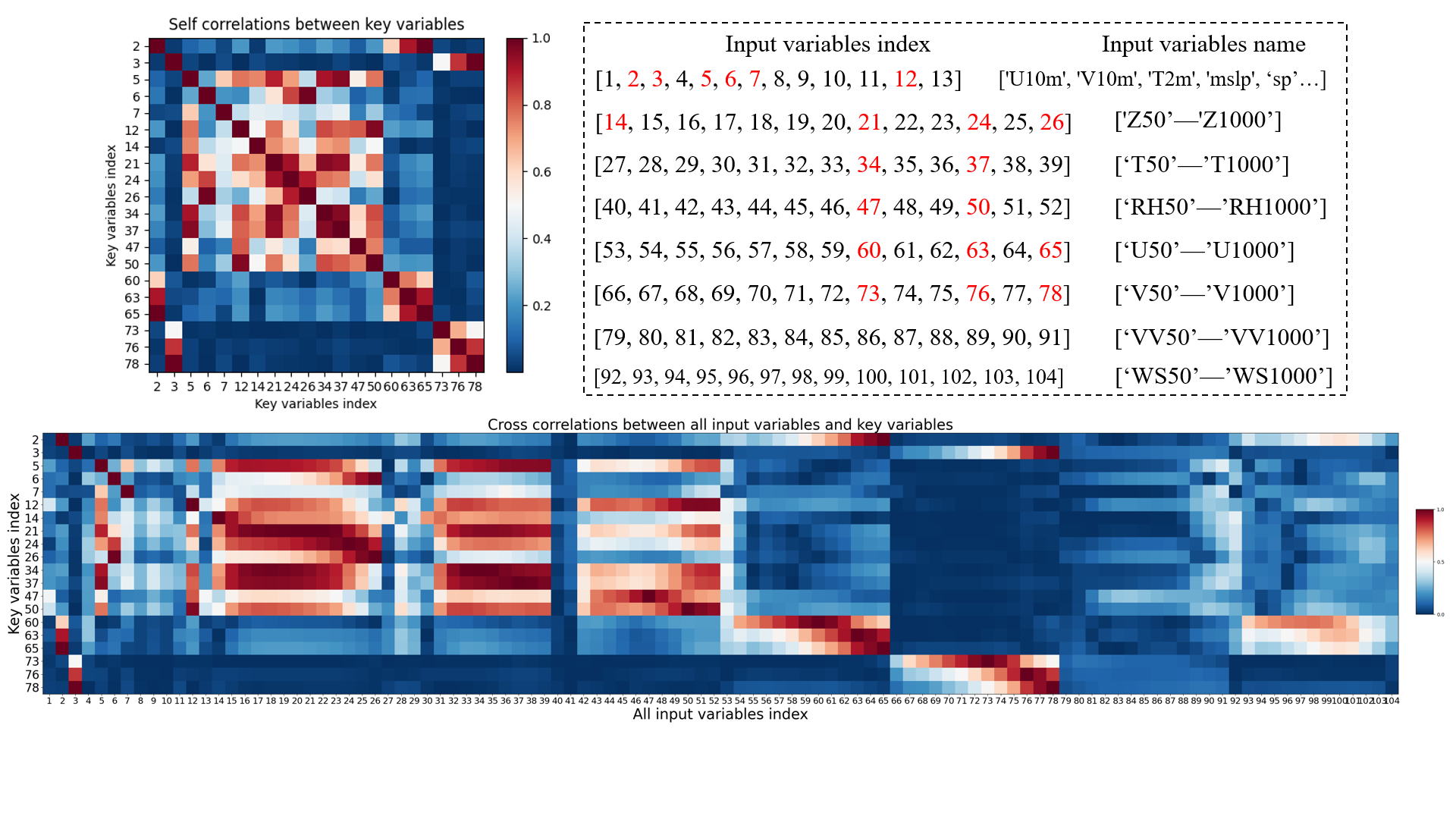}
    \caption{The self correlation coefficient between key variables and cross correlation between key variables and all input variables of ERA5 dataset. The closer it is to 1, the higher correlation and the closer it is to 0, the lower correlation. We also list all input variables and the index with red color represent the key variables. It could be seen that 1: the self correlations between key variables is mostly weak. 2: almost every key variable is strongly correlated with several certain atmospheric variables of continuous pressure levels.
}
    \label{fig:corr_mat}
\end{figure*}

In recent years, deep learning methods have been increasingly applied to the field of weather forecasting due to their strong capability to learn the evolutionary patterns of features and lower requirement of computation resources and short time in inference stage \cite{https://doi.org/10.1029/2019MS001705}\cite{https://doi.org/10.1029/2020MS002109}. 
As shown in below left of Fig. \ref{fig:overall_frame}, some methods adopt the convolutional neural network (CNN), graph neural networks (GNN), Adaptive Fourier Neural Operators (AFNO) \cite{guibas2022adaptive} or transformer \cite{NIPS2017_3f5ee243} network to perform iterative prediction in the original variable space to obtain predicted results for multiple lead times \cite{hua2024weather}\cite{pathak2022fourcastnet}\cite{10475377}. However, predicting key variables and learning their evolutionary patterns in their original variable space is quite challenging \cite{https://doi.org/10.1029/2020MS002405}\cite{hua2024weather}. Because there are intricate interactions between key variables and other atmospheric variables, resulting in weak correlations of the evolutionary patterns among only key variables, as shown in top of Fig. \ref{fig:corr_mat}. 



As shown in Fig. \ref{fig:corr_mat}, the key variables are strongly correlated with several certain atmospheric variables of continuous pressure levels. So input these related variables is expected to more accurately capture the evolutionary patterns of key variables by fully excavating
and learning the intricate interaction among key variables and other related variables   \cite{https://doi.org/10.1029/2022MS003211}\cite{https://doi.org/10.1029/2020MS002405}. And some studies have incorporated more variables to improve the prediction performance of key variables. For example, Weyn et al. \cite{https://doi.org/10.1029/2019MS001705} constructed the CNN and LSTM prediction model to predict the Z500 (geopotential at 500 hPa) and experimentally verified that adding the 700- to 300-hPa thickness as input could improve the prediction performance. 
Hu et al. \cite{https://doi.org/10.1029/2022MS003211} choose four key variables as the prediction targets. And considering the complex interactions between different atmospheric variables, they also choose 67 additional variables to provide additional feature information for the targets prediction. Nevertheless, to effectively incorporate more variables and learn their evolving patterns across different lead times, these methods adopt an iterative prediction manner. In each iterative step, the model must predicts all input variables for the next time step \cite{https://doi.org/10.1029/2019MS001705}\cite{https://doi.org/10.1029/2020MS002109}, as shown in bottom left of Fig. \ref{fig:overall_frame}. This significantly increases the difficulty of the model in learning the evolving patterns of a large number of variables at different time steps. Additionally, the evolutionary patterns and data distributions of different variables are inconsistent, making it difficult for neural networks to learn simultaneously and consuming substantial computational power \cite{bi2023accurate}\cite{chen2023fuxi}\cite{doi:10.1126/science.adi2336}\cite{pathak2022fourcastnet}\cite{doi:10.1126/science.adi2336}\cite{chen2023fengwupushingskillfulglobal}. Therefore, it is crucial to efficiently and effectively incorporate more variables to more accurately excavate and learn the evolutionary patterns of key variables.


Hence, this paper propose a \textbf{k}ey \textbf{v}ariable \textbf{p}rediction model based on \textbf{l}atent \textbf{s}pace \textbf{i}teration (Kvp-lsi) with inputting more variables, as shown in right part of Fig. \ref{fig:overall_frame}. To capture the complex evolutionary patterns of key variables, we first extract the low-dimensional latent features that significantly correlated with key variables from all input variables, and then feeds the obtained latent features into the predictor to iteratively forecast latent features for multiple lead times. The predicted multi-steps latent features are then decoded in parallel to obtain predicted key variables at the corresponding lead time. In addition, during the training stage, guidance for the latent feature prediction is provided, namely the loss function between the encoded latent features and the predicted latent features at the corresponding lead time, to improve the long-term iterative accuracy of latent feature. To further enhance the long-term prediction accuracy, we design the hierarchical temporal aggregation with more inputs (HTAMI), an algorithm that minimizes the number of model iterations when predicting multiply long lead time, effectively mitigating cumulative errors. This algorithm, which inputs more (two) time steps rather than one, is an improved version of \cite{bi2023accurate} and further enhancing the temporal correlation between the predicted and input variables.



\section{Related Work}\label{sec_Related}

To effectively enhance the temporal correlation between input and predicted results, Recurrent Neural Networks (RNN) \cite{zaytar2016sequence} and Long Short-Term Memory (LSTM) \cite{SALMAN201889} models have been applied in the weather forecasting field. These types of models perform the iterative prediction and incorporate the input variable at current time step and past hidden state to predict hidden state at next time step. For example, Shi \cite{NIPS2015_07563a3f} designed the ConvLSTM model which combine the CNN and LSTM and achieved the superior performance on Moving-mnist and radar echo datasets. Then, Shi et al. \cite{NIPS2017_a6db4ed0} constructed the TrajGRU network to learn the location-variant structure and showed better performance. However, these models suffer from several problems. For instance, estimating hidden state $h_t$ in time $t$ requires incorporating $h_{t-1}$ and input variable $x_{t-1}$ which introduce more error sources and consequently amplifies the error in estimating $x_t$ from the resulted $h_t$, especially in multi-round iterations. Moreover, the initialization method of hidden state can also deteriorate the prediction performance. 

Inspired by the NWP framework, some methods have been developed to perform iterative predictions within the original atmospheric variable space. For example, Weyn et al. \cite{https://doi.org/10.1029/2020MS002109} proposed the CNN-based iterative prediction network in the cubed-sphere that output two time steps for each iteration. 
With the increase in GPU computing power, some studies have introduced more atmospheric variables to fully learn the interactions between different atmospheric variables to improve the accuracy of iterative predictions. For example, Pathak et al. \cite{pathak2022fourcastnet} adopted the pre-training and finetune strategy and predicted 20 upper-air and surface variables, which match the accuracy of ECMWF's IFS. Bi et al. \cite{bi2023accurate} introduced more upper-air in 13 vertical levels, and constructed 3D earth-specific transformer-based network and hierarchical temporal aggregation (HTA) algorithm to improve the long-range prediction accuracy, which achieved the better prediction performance than the world's best NWP system. Lam et al. \cite{doi:10.1126/science.adi2336} designed the GNN-based iterative prediction network, which outperforms the most accurate weather forecasting system--HRES in almost 224 atmosphere variables. Chen et al. \cite{chen2023fuxi} adopted the Unet-transformer network and fine-tuned three prediction models for different forecasting ranges. However, it is difficult for neural networks to learn the complex evolutionary patterns and inconsistent data distributions of a large number of atmospheric variables, which consumes significantly large computational power.

\section{Method}\label{sec_method}
\subsection{Dataset}

\begin{table*}[h]
\centering
\caption{The abbreviations of the key variables. U10m and V10m represent the zonal and meridonal wind velocity at the height of 10m above the surface; T2m represents
the temperature at 2m above the surface; T, U, V, Z and RH represent the temperature, zonal wind velocity, meridonal wind velocity, geopotential and relative humidity at specified
vertical level; TCWV represents the total column water vapor. The numbers in parentheses represent the index of key variables in all input variables as in Fig. \ref{fig:corr_mat}.}\label{tab_key_var}%
\begin{tabular}{cc}
\hline
Vertical level & Variables \\
\hline
Surface  & U10m(2), V10m(3), T2m(5), sp(7), mslp(6)   \\
1000 hPa & U(65), V(78), Z(26)  \\
850 hPa & T(37), U(63), V(76), Z(24), RH(50)   \\
500 hPa & T(34), U(60), V(73), Z(21), RH(47)  \\
50 hPa & Z(14)   \\
Integrated & TCWV(12)  \\
\hline
\end{tabular}
\end{table*}

The validity of the proposed prediction model was verified through experiments conducted on the ERA5 dataset from Weatherbench2 \cite{rasp2024weatherbench}. Due to memory constraints, we selected the global dataset from Weatherbench2 with a spatial resolution of 5.625°. Its temporal resolution was downsampled to 6 hour, and the key atmospheric variables for prediction included multiple upper-air and surface variables as listed in Table \ref{tab_key_var}, with a total forecasting period of 15 days (60 lead time steps). Note that these key variables are chosen following the \cite{pathak2022fourcastnet}\cite{gan2024ewmoe}. And these variables are pretty important to support predicting severe events, including tropical cyclones, atmospheric rivers and extreme temperature \cite{bi2023accurate}\cite{doi:10.1126/science.adi2336}.

To more accurately capture the evolutionary patterns of key variables, we input more atmospheric variables than key variables, including 'geopotential', 'temperature', 'specific humidity', 'u component of wind', 'v component of wind', 'vertical velocity' and 'wind speed' at 13 different vertical levels. Additionally, 13 surface variables were included. So the input includes 104 variables, with the key variable being among them. Constant variables and time embeddings were also incorporated. As the data ranges of different variables varied, each variable was normalized before entering the model – that is, mean-subtracted and divided by its standard deviation. The predicted key variables of model were then reversely normalized. We select the ERA5 dataset from 2000 to 2020 year to conduct the experiments. And the datasets for model training, validation, and testing were segmented into the periods of 2000-2017, 2018-2019, and 2020, respectively. 



\subsection{Overall framework}

\begin{figure*}[t]
\centering
    \includegraphics[width=1\textwidth,trim=60 110 45 0,clip]{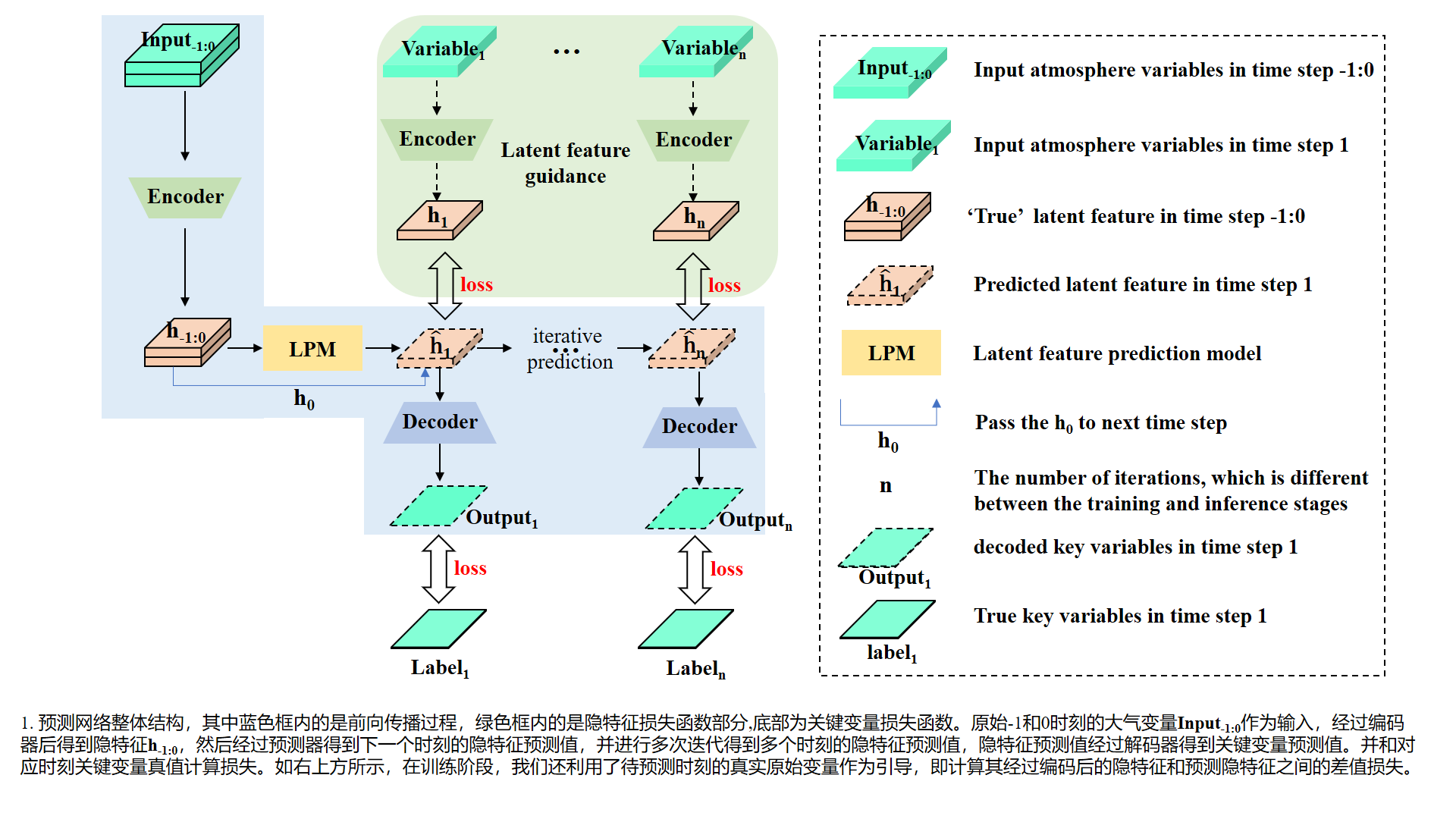}
    \caption{The overall prediction model structure. The blue box indicates the forward propagation process ‘encode-predict-decode’. The green box contains the latent feature loss function component, and the bottom part represents the key variable loss function. 
    }
    \label{fig:network}
\end{figure*}

To effectively learn complex temporal evolutionary patterns of key variables, we encode the input multiple atmospheric variables into the inter-correlated latent features, which is significantly related to the key variables. It is easier to capture the evolutionary patterns of these latent features than original key variables.

Therefore, we construct an encode-predict-decode network, as shown in Fig. \ref{fig:network}. Initially, we construct an encoder to extract the low-dimensional latent features strongly correlated with the key variables from the inputs of much more atmospheric variables than key variables. Subsequently, the predictor iteratively forecasts these latent features, guided by the original atmospheric variables at the corresponding lead time. At last, the predicted atmospheric latent features obtained from iteration are then fed into a decoder to yield the forecasting values of the key variables at the corresponding lead time.

It is noteworthy that the proposed network is fed with two time steps and predict the latent feature of the next single time step. For longer-range forecasts, we unfold the model by iteratively feeding its predictions alongside the previous time step's latent feature back to the model as input. The time interval between two time steps is 6 hours, and for the sake of clarity, we use different subscripts to denote different time steps, as shown in Fig. \ref{fig:network}.

\subsection{Weather forcasting model}
\subsubsection{Encoder}

\begin{figure*}[t]
\centering
    \includegraphics[width=1\textwidth,trim=0 140 80 0,clip]{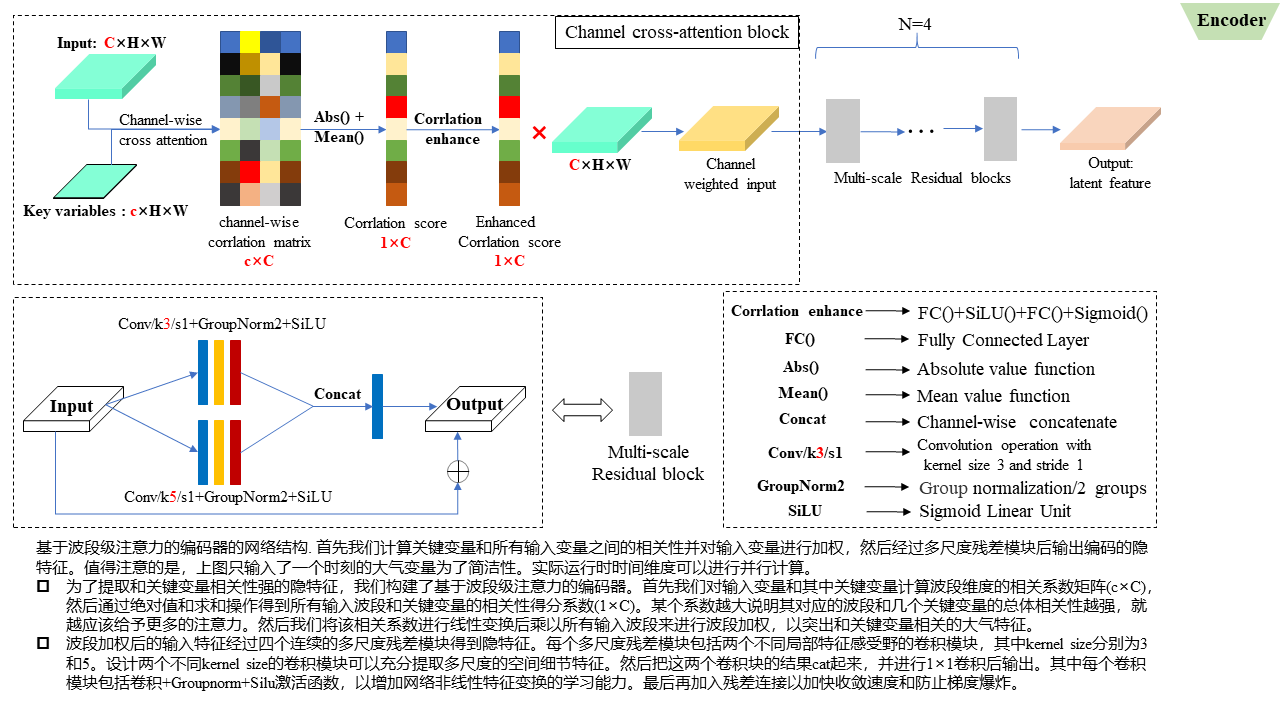}
    \caption{Network structure of the channel-wise cross-attention based encoder. 
    It is worth noting that the above figure only shows the atmospheric variables of one time step for simplicity. In actual operation, different time steps can be calculated in parallel.
}
    \label{fig:encoder}
\end{figure*}

To extract low-dimensional latent features strongly correlated with key variables, we construct an encoder based on the channel-wise attention, as shown in Fig. \ref{fig:encoder}. Initially, we calculate the correlation coefficient matrix in the channel dimension for the input variables and the key variables among them. Then, through absolute, summation and linear transform operations, the resulted correlation score are multiplied by the input variables to reweight different variables.

After weighting different variables, the input features are passed through four consecutive multi-scale residual blocks to obtain the latent features. As shown in Fig. \ref{fig:encoder}, each multi-scale residual block consists of two convolutional modules with the kernel sizes of 3 and 5, respectively. Designing convolutional modules with two different kernel sizes allows for the full extraction of multi-scale spatial features.

Note that the input atmosphere variables with two time steps are processed parallel in time dimension, which is as detailed below, the encoder would receives two time steps with varying intervals for different lead times and receives five time steps in inference stage. In this research, the number of channels of latent feature is set to 24 in this research, which is much lower than that of the original input variables.

\subsubsection{Prediction model}

\begin{figure*}[t]
\centering
    \includegraphics[width=1\textwidth,trim=20 290 380 0,clip]{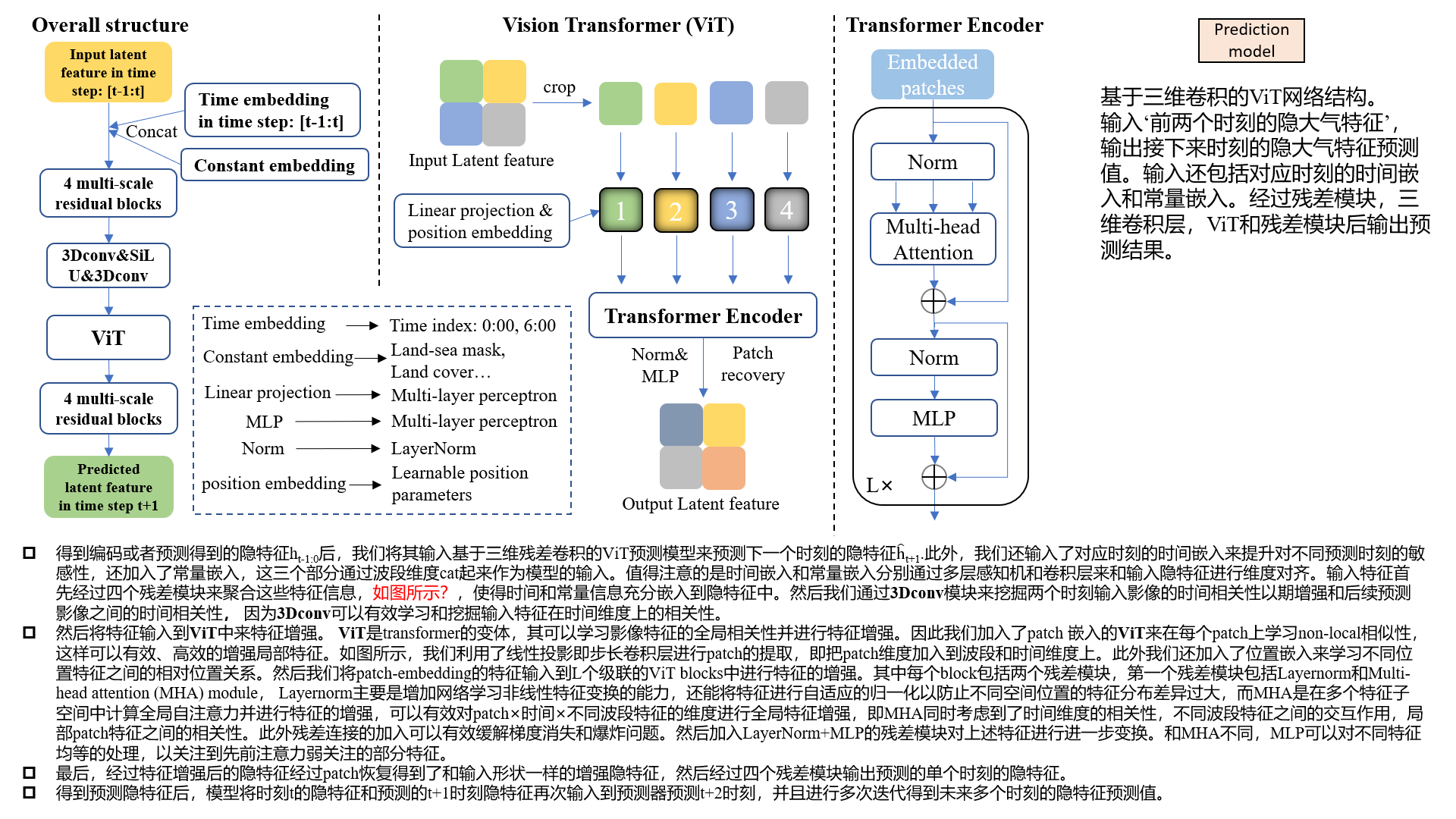}
    \caption{The network structure of 3d-ViT based latent feature prediction model (LPM). 
}
    \label{fig:LPM_network}
\end{figure*}

After obtaining the encoded latent features $ h_{-1:0} $ (this subscript denotes
two time steps -1 and 0), we input them into the 3-dimensional convolution and vision transformer (3d-ViT) based prediction model to predict the next time step's latent feature $\hat{h}_{1}$. Additionally, we include the corresponding time embedding to enhance the sensitivity to different input times and add the constant embedding. These three components are concatenated along the channel dimension to serve as the model input. 

The input features first pass through four multi-scale residual blocks (as in Fig. \ref{fig:encoder}) to aggregate different feature information, as shown in the left part of Fig. \ref{fig:LPM_network}, allowing time and constant information to be fully embedded into the latent features. We then employ a 3-dimensional convolution (3Dconv) module to mine the temporal correlation between two time steps’ input latent features, aiming to enhance the temporal correlation with the predicted latent feature.

The latent features $h_{2d\times hw}$ are then fed into ViT for feature enhancement. ViT \cite{dosovitskiy2020image} is a variant of the transformer \cite{NIPS2017_3f5ee243} that can learn the global correlation of image features and enhance them. And it introduce the patch embedding to efficiently learn the patch-wise non-local attention to enhance the patch features. Finally, enhanced latent features are passing through four multi-scale residual blocks, to output the predicted latent feature in next time step.  

Upon obtaining the predicted latent features, the LPM takes the latent features of time step t and the predicted latent features of time step t+1 as input and forecast the latent features of lead time step t+2, and iterates multiple rounds to obtain multiple lead time steps' latent features.

\subsubsection{Decoder}

The multiple lead time steps' latent atmospheric features generated by iterative forecasting, provide the predicted key variables after decoding. The decoder network architecture consists of four multi-scale residual blocks, as illustrated in Fig. \ref{fig:encoder}.

\subsection{Model iterative manner}

To mitigate the cumulative errors, \cite{bi2023accurate} proposed the HTA algorithm to minimize the number of prediction model iterations, that is, by training multiple models with different time intervals. However, for this algorithm, each predictive model only takes the previous single time step as input to predict the next time step, making it difficult for the model to capture the complex temporal dependencies between the input and output. And the HTA algorithm in \cite{bi2023accurate} cannot directly input two time steps to predict the next one due to the constraint of this algorithm.

\begin{figure*}[t]
\centering
    \includegraphics[width=1\textwidth,trim=50 310 50 70,clip]{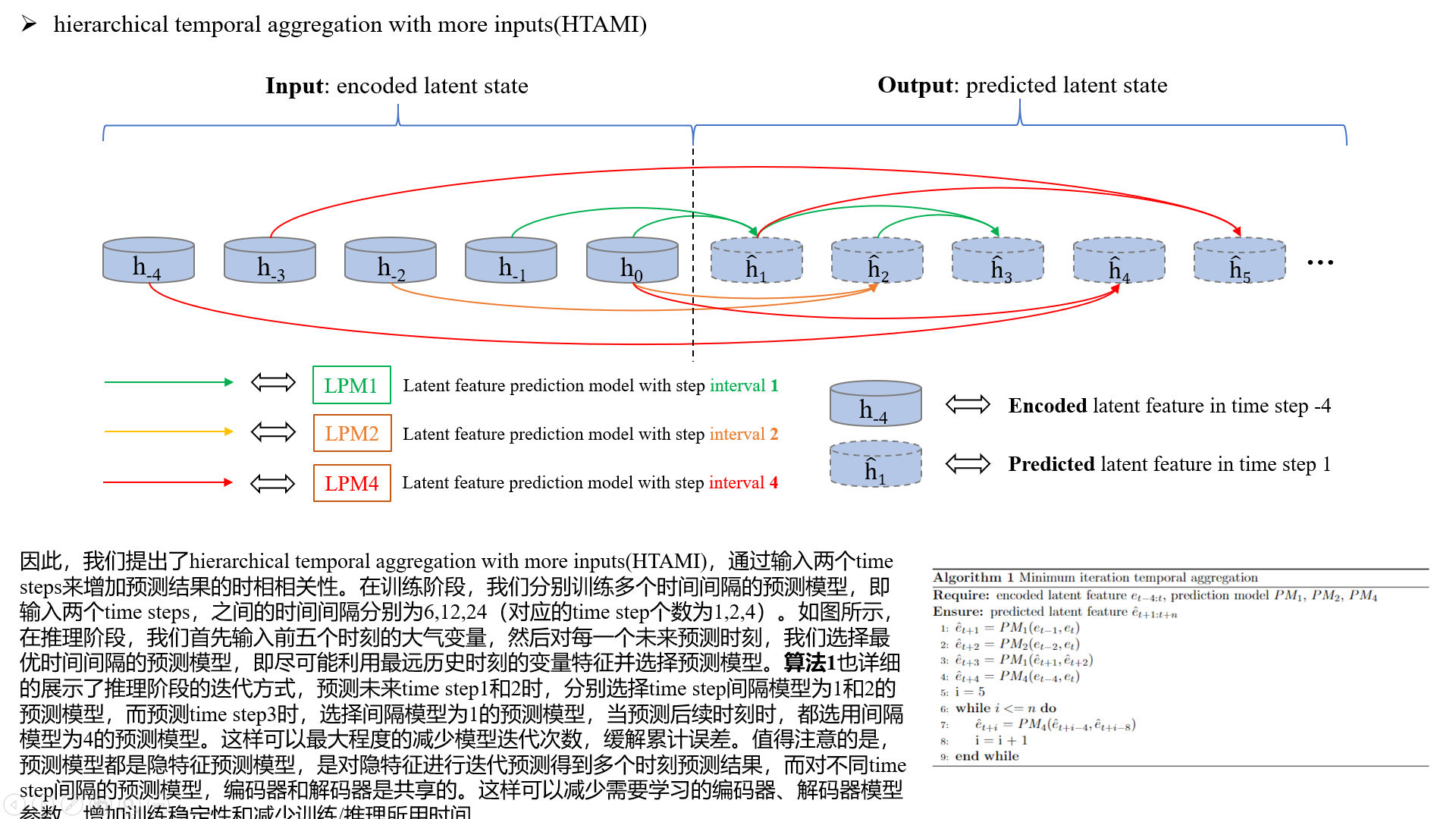}
    \caption{The schematic of the proposed HTAMI. During the inference stage, the model accepts five time steps as input and selects the longest interval prediction model for each lead time to ensure the least total number of iterations. 
}
    \label{fig:iteration_manner}
\end{figure*}

\begin{algorithm}
\caption{Hierarchical temporal aggregation with more inputs (HTAMI) }\label{iteration_algo1}
\begin{algorithmic}[1]
\Require encoded latent feature $h_{-4:0}$, prediction model $LPM1$, $LPM2$, $LPM4$
\Ensure predicted latent feature $h_{1:n} (n\geq 4)$ 
\State $h_{1}$ = $LPM1(h_{-1}, h_{0})$
\State $h_{2}$ = $LPM2(h_{-2}, h_{0})$
\State $h_{3}$ = $LPM1(h_{1}, h_{2})$

\State i = 4
\While{$ i <= n $}
    \State $h_{i}$ = $LPM4(h_{i-4}, h_{i-8})$
    \State i = i + 1
\EndWhile
\end{algorithmic}
\end{algorithm}

Therefore, we propose Hierarchical Temporal Aggregation with More Inputs (HTAMI) algorithm, which increases the temporal correlation of the predicted results by taking two time steps as input other than one. During the training stage, we train multiple LPMs with different time intervals, that is, taking two time steps with intervals of 6, 12, and 24 hours as inputs and predict the next time step with the corresponding time interval. The above time intervals are equal to 1, 2, and 4 time steps and corresponding to LPM1, LPM2 and LPM4, as shown in Fig. \ref{fig:iteration_manner}, respectively. During the inference stage, as depicted in Fig. \ref{fig:iteration_manner}, we first input the atmospheric variables from the previous five time steps. Then, for each lead time step, we select the optimal time interval prediction model. Algorithm \ref{iteration_algo1} details the iterative procedure during the inference stage. 

This strategy effectively enhance the temporal correlation between the predicted results and input. It is noteworthy that the prediction models are latent feature prediction models (LPM), which means the overall prediction network include one encoder/decoder and 3 LPMs with different prediction time intervals. And in the training stage we randomly choose one LPM to optimize in each training step, while in inference stage we select different LPMs to effectively mitigate the cumulative errors. This can greatly reduce the number of encoder and decoder model parameters that need to be learned and reducing the computation cost in the training and inference stages.

\subsection{Loss functions}

\begin{align}
\mathcal{L}_{\mathrm{L1\_key}}=\sum_{d_0\in D_{\mathrm{batch}}}
\underbrace{\sum_{\tau\in1:T_{\mathrm{train}}}}_{\text{lead time}}\underbrace{\sum_{j\in J}}_{\text{variables}} q_jw_ja_i\underbrace{|\hat{k}_{i,j}^{d_0+\tau}-k_{i,j}^{d_0+\tau}|}_{\text{absolute error}}
\label{MSE_key}
\end{align}

\begin{align}
\mathcal{L}_{\mathrm{L1\_recons}}=\sum_{d_0\in D_{\mathrm{batch}}}\underbrace{\sum_{j\in J}}_{\text{variables}}
a_i\underbrace{|\hat{k}_{i}^{d_0}-k_{i}^{d_0}|}_{\text{absolute error}}
\label{MSE_recons}
\end{align}

$\bullet D_0\in D_{batch} \text{ represent forecast initialization date-times in a batch of forecasts in } \\ \text{ the training set}. $

$\bullet\tau\in1:T_{\mathrm{train}}\text{ are the lead time steps of the predictor model in the training stage}.$

$\bullet j \in J \text{ indicates different variables of which weights $w_j$ for surface variables are 0.1} \\ \text{except the 'T2m' is 1 and upper-air variables are 1}.$


$\bullet q_j \text{ is the per-variable inverse variance}.$

$\bullet a_i \text{ is the weight which varies with latitude, and is normalized to
unit mean}. $ 

To effectively optimize the network parameters, we constructed three L1 loss functions. L1 loss could alleviate the blurring effect greatly. The first loss function Eq. \ref{MSE_key} is between the key variables decoded from the predicted latent features and the true values in ERA5. Moreover, different dimensions of the key variables are weighted as shown in the Eq. \ref{MSE_key}, including the inverse variance per variable, distinct weights for surface and atmospheric variables, and varying weights for different latitudes. The inverse variances are designed to reduce the distributional differences among different variables in different times. Subsequently, the weights for surface variables and atmospheric variables are set to 0.1 and 1 respectively except for T2m of 1 \cite{bi2023accurate}. Furthermore, since the actual areas corresponding to different grids vary, a cosine decay scheme was applied to weight the grids at different latitudes.

In addition, we incorporate a reconstruction loss, which is calculated by decoding the latent feature that is directly encoded from the original input variables at all time steps (include the input and output), and then computing the loss against their true values. This loss further guides the update of the encoder and decoder parameters, preventing them from falling into local minima.

\begin{align}
\mathcal{L}_{\mathrm{L1\_latent}}=\sum_{d_0\in D_{\mathrm{batch}}}
\underbrace{\sum_{\tau\in1:T_{\mathrm{train}}}}_{\text{iteration steps}}a_i\underbrace{|\hat{h}_{i}^{d_0+\tau}-h_{i}^{d_0+\tau}|}_{\text{absolute error}}
\label{MSE_latent}
\end{align}

To improve the accuracy of the predicted latent feature in multi-round iterations, we construct the latent feature loss function to guide the predicted latent feature close to the encoded latent feature. For this loss function, only the latitude and the lead time are being weighted. The motivation of this loss can be attributed to the fact that, following the above reconstruction loss, the encoded latent features, once processed by the decoder, closely resemble the true values of the key variables at the corresponding lead times.

Therefore, the overall loss function are the combination of above three loss functions:

\begin{align}
\mathcal{L}_{\mathrm{overall}}= \mathcal{L}_{\mathrm{L1\_key}} + \mathcal{L}_{\mathrm{L1\_recons}} + \mathcal{L}_{\mathrm{L1\_latent}}
\label{MSE_all}
\end{align}

\noindent note that the weights of these three loss functions are identical.

\section{Experiment}

\subsection{Experimental details}

The number of training epochs is set to 65. During training, we use a curriculum schedule with four stages, which set the number of iterations for the LPM gradually increased with epochs, with an iterations number schedule of [2,4,6,8], corresponding to the epochs of 0-50, 50-55, 55-60, and 60-65, respectively. The initial learning rate was set to 2e-4, which was multiplied by 0.5 every 10 epochs during the 0-50 epoch period, and remain unchanged thereafter. The optimizer used is Adam with the parameters $\beta_1$=0.9, $\beta_2$=0.95 and a high weight decay of 1. The batch size is set to 32. All experiments are run based on the PyTorch framework with two NVIDIA RTX 3090 GPUs.

\subsection{Compared methods and metric index}

To validate the effectiveness of the proposed method, we select several state-of-the-art (SOTA) methods for comparison, including PredRNNv2 \cite{9749915}, SimVP \cite{Gao_2022_CVPR} and FourCastNet \cite{pathak2022fourcastnet}. To take a more fair comparsion, we also add the HTA algorithm to SimVP and FourCastNet methods, which are called SimVP\_HTA and FourCastNet\_HTA. Notably, we reproduce all above methods on our device. 

This study employ two performance metrics to measure the consistency between the predicted key variables and their true values in ERA5, including Root Mean Square Error (RMSE) and anomaly correlation coefficient (ACC). And both indices are weighted across different latitudes. A lower RMSE index and higher ACC index indicate more accurate prediction results.

\subsection{Comparative results with other SOTA methods on global dataset}

\begin{figure*}[h]
\centering
    \includegraphics[width=1\textwidth,trim=0 0 0 0,clip]{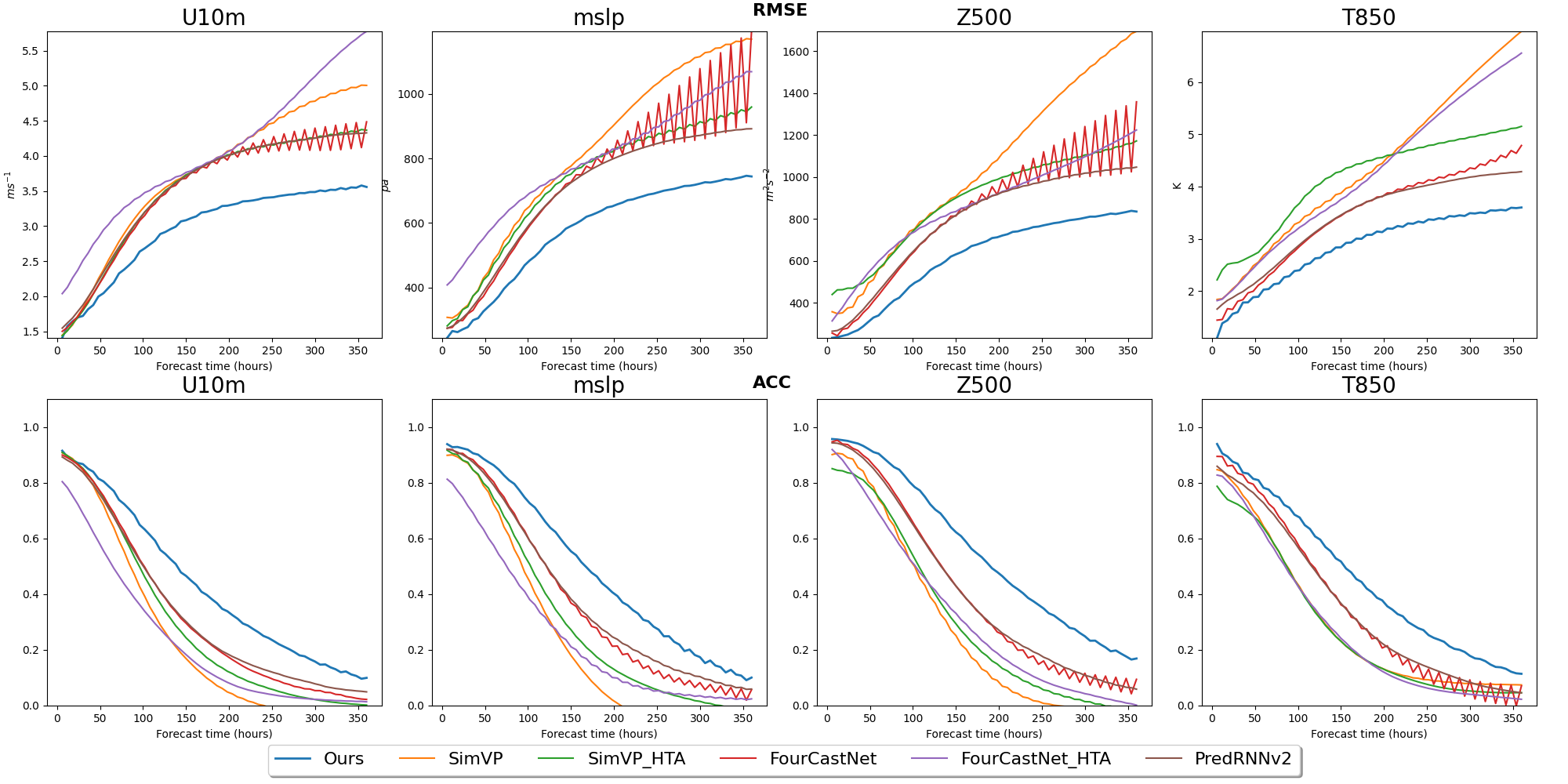}
    \caption{The averaged quantitative comparison results on the global 5.625° resolution dataset. We display several key variables include zonal wind velocity at 10m surface height (U10m), mean sea level pressure (mslp), geopotential on 500hpa (Z500) and temperature on 850hpa (T850).}
    \label{fig:metric_U10m}
\end{figure*}

Fig. \ref{fig:metric_U10m} presents the quantitative comparison results of different methods. It is evident that, our proposed method significantly outperforms all other methods in terms of short-term, mid-term, and long-term RMSE and ACC indices across displayed four key variables. And the gap in RMSE metrics between our method and others widens over lead time, which fully demonstrates the comprehensive advantage of our approach across all forecasting periods. By comparing the outcomes of the SimVP and SimVP\_HTA methods, it can be inferred that the integration of the HTA algorithm significantly enhances the performance of mid-to-long term predictions. However, the introduction of this algorithm leads to a degradation in short-term performance, as shown in last two columns of Fig. \ref{fig:metric_U10m}. Moreover, the HTA algorithm can strengthen the temporal correlation of the forecasted outcomes, as evidenced by the long-term results of the FourCastNet and FourCastNet\_HTA methods.

\begin{figure*}[h]
\centering
    \includegraphics[width=0.8\textwidth,trim=40 60 0 70,clip]{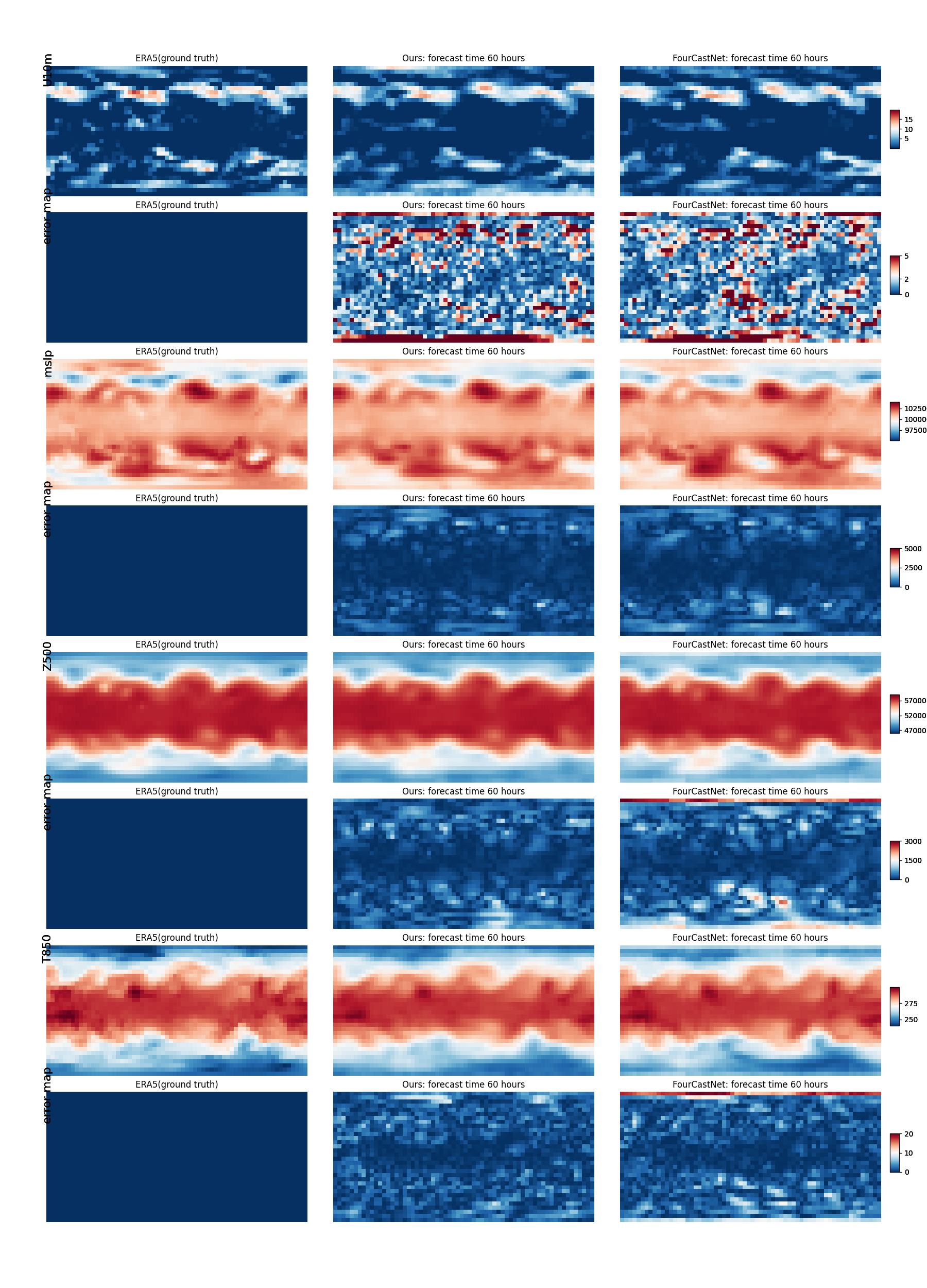}
    \caption{The predicted results of lead time 60 hours that take 5 time steps before 2020.02.25.12:00 as the model input. Different rows display different key variables, which include U10m, mslp, Z500 and T850. We also present error maps on even rows.
}
    \label{fig:quali_fig_U10m}
\end{figure*}

We also showcase the qualitative forecast results and error maps, as shown in Fig. \ref{fig:quali_fig_U10m}. Due to page limitations, we choose FourCastNet, which had the sub-optimal predictive performance compared to our method, for comparison. It is evident that the predicted results of our method are closer to the true values compared to FourCastNet. For example, as shown by third row in Fig. \ref{fig:quali_fig_U10m}, our method more accurately predict the range and intensity of the low-pressure center of mslp in the South Pacific. Our method also made more precise predictions regarding the Z500 and T850 intensity in the South Pacific. Above quantitative and qualitative forecast results comparisons indicate that our method can more accurately learn the evolutionary patterns of key variables.

\subsection{Comparative results with other SOTA methods on longer lead times}

\begin{figure*}[h]
\centering

\includegraphics[width=0.8\textwidth,trim=40 60 0 70,clip]{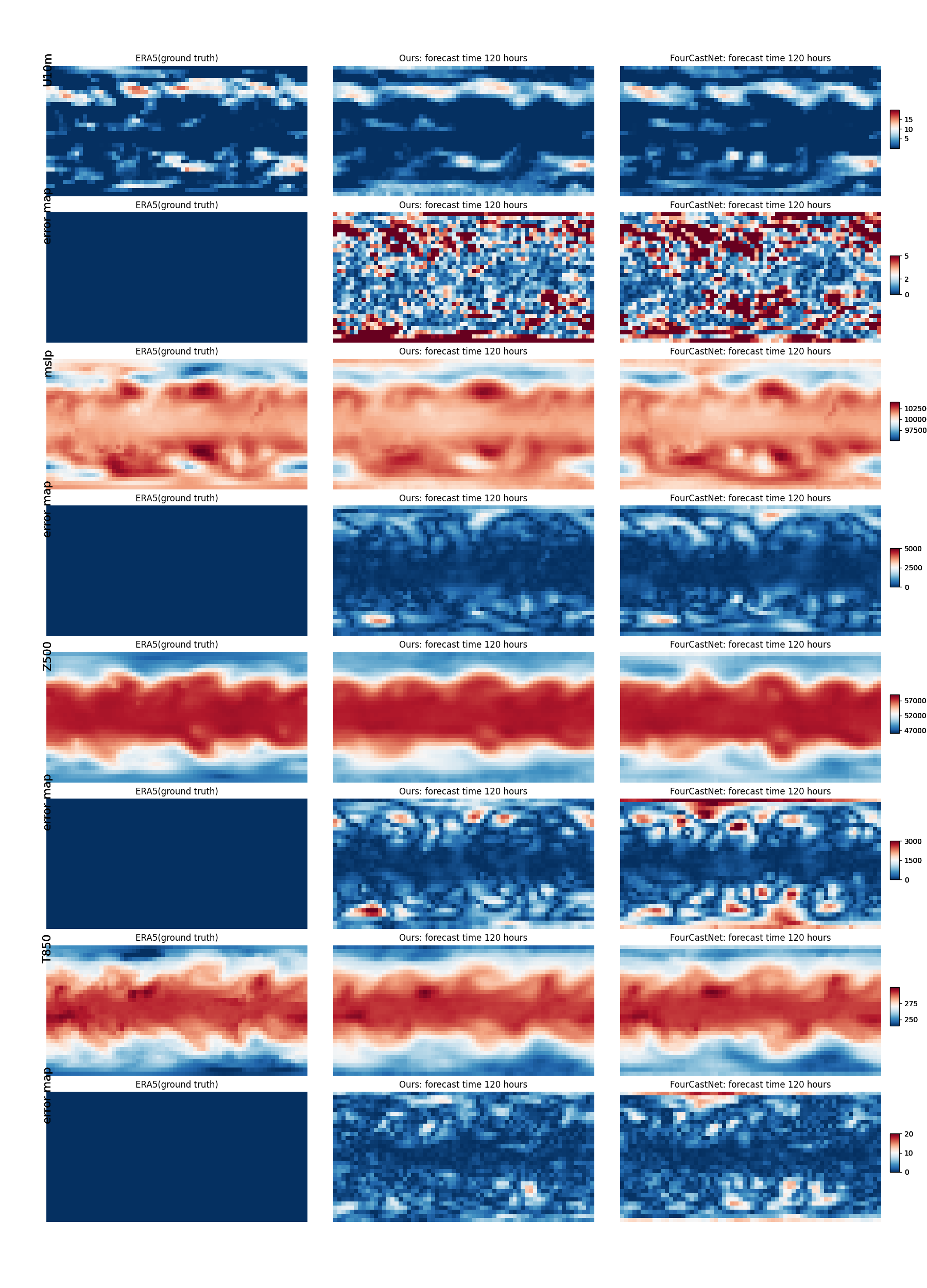}
    \caption{The predicted results of lead time 120 hours that take 5 time steps before 2020.02.25.12:00 as the model input. The key variables displayed include U10m, mslp, Z500 and T850.}
    \label{fig:metric_mslp3}
\end{figure*}

\begin{figure*}[h]
\centering

\includegraphics[width=0.8\textwidth,trim=40 60 0 70,clip]{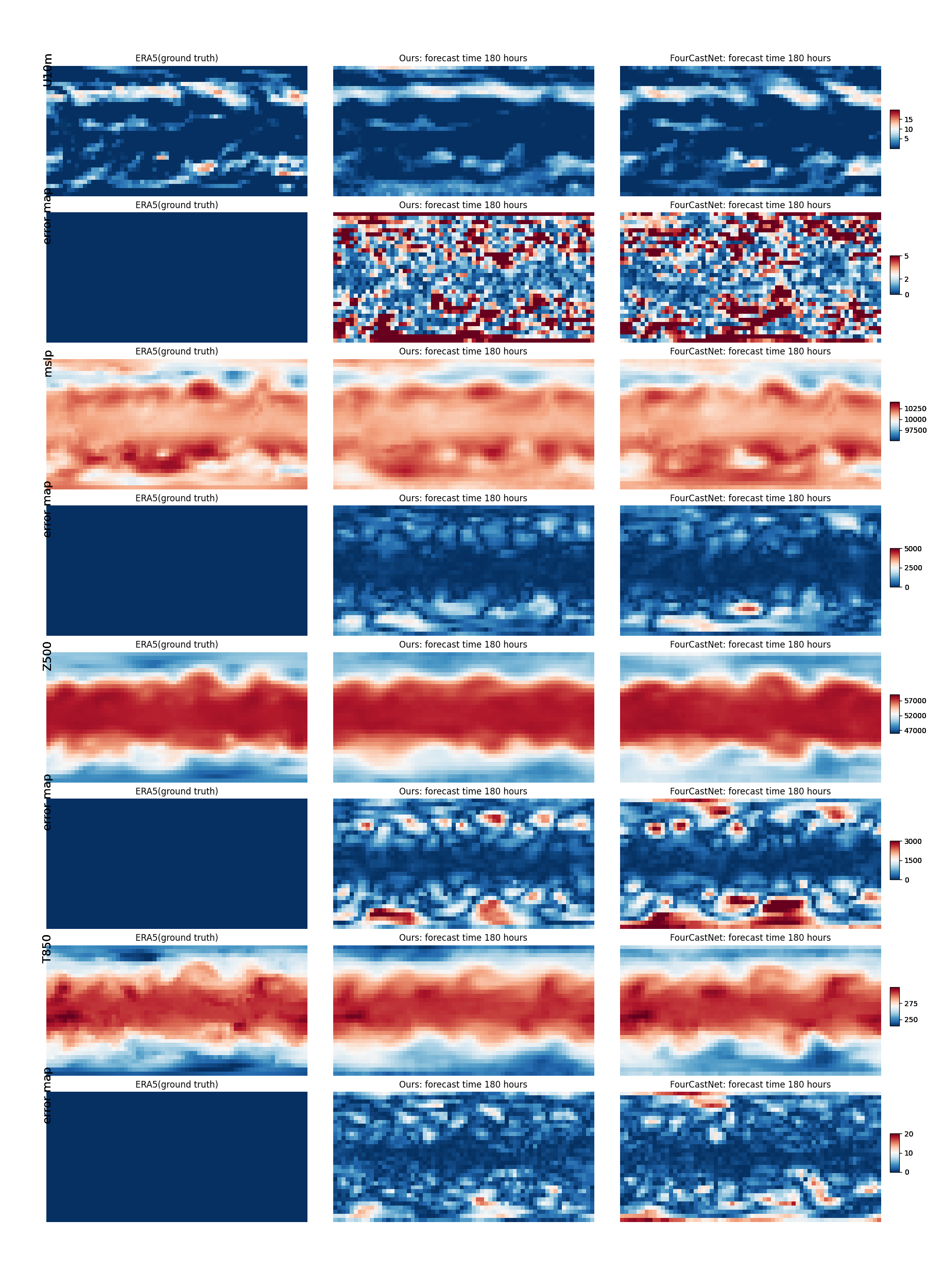}
    \caption{The predicted results of lead time 180 hours that take 5 time steps before 2020.02.25.12:00 as the model input. The key variables displayed include U10m, mslp, Z500 and T850.}
    \label{fig:metric_mslp4}
\end{figure*}

We also present qualitative forecasts results for longer lead times to demonstrate the advantages of our method, including lead times of 120 and 180 hours. As shown in Fig. \ref{fig:metric_mslp3} and \ref{fig:metric_mslp4}, it can be inferred that our method achieves better prediction performance than sub-optimal method--FourCastNet at longer lead times. 
For example, in the 120-hour forecast results, our method more accurately predict the low-pressure areas of mslp near the North Pole, as shown in the fourth row of Fig. \ref{fig:metric_mslp3}. For Z500, our method also more accurately predict the trend of low-pressure changes around the North and South poles. For temperature at 850 hPa (T850), the temperature trend changes in the South Pacific and the Arctic are also more accurately captured by our method, as shown in the last two rows of Fig. \ref{fig:metric_mslp3}. As the lead time increases, the prediction results of all methods deteriorate, but our method's predictions still outperform those of FourCastNet, as shown in Fig. \ref{fig:metric_mslp4}. The prediction error of FourCastNet is very large, especially for Z500 and T500, which can be clearly seen from the error maps of Fig. \ref{fig:metric_mslp4}. In summary, our method achieves the best predictive performance across all different lead times.

\section{Discussion}

\subsection{The manner to predict multiple lead times}

Some method adopt different manners to predict multiple lead times. For example, \cite{https://doi.org/10.1029/2020MS002109}\cite{10475377} predicts multiple lead times at a single output, the training difficulty increases, and results in a weaker temporal correlation between the predicted results at adjacent lead times. Additional, when the prediction model receives an 'lead time embedding' to forecast the result at that lead time, such as \cite{espeholt2022deep}, the results for different lead times do not depend on previous time steps, making it more challenging for the network to capture their evolutionary patterns (requiring more data and computational resource). Therefore, it is desirable to adopt the iterative prediction method and output one time step in each iteration, which can make it easier for the network to capture their evolutionary patterns and greatly enhance the temporal correlation between the input and output sequence.

For the iterative prediction method, several strategies have been designed to mitigate the accumulation errors of multi-round iteration. For instance, Bi et al. \cite{bi2023accurate} designed the HTA algorithm, which minimizes the number of iterations by training multiple models with different prediction intervals. Chen et al. \cite{chen2023fuxi} initially pre-trained a predictive model and then fine-tuned it to obtain three predictive models with different lead time ranges. Given that separately fine-tuning for each predictive model consumes substantial computational power, we adopted the HTA algorithm to reduce error accumulation. And we enhanced the HTA algorithm by inputting more time steps to strengthen the temporal correlation between predictions and inputs. Note that we training multiple prediction models with varying time intervals in the latent space with the shared encoder and decoder, which significantly reducing computational costs.

\subsection{Ablation study}

\subsubsection{The effect of latent space iterative prediction and HTAMI }

\begin{figure*}[t]
\centering
\includegraphics[width=1\textwidth,trim=0 0 0 0,clip]{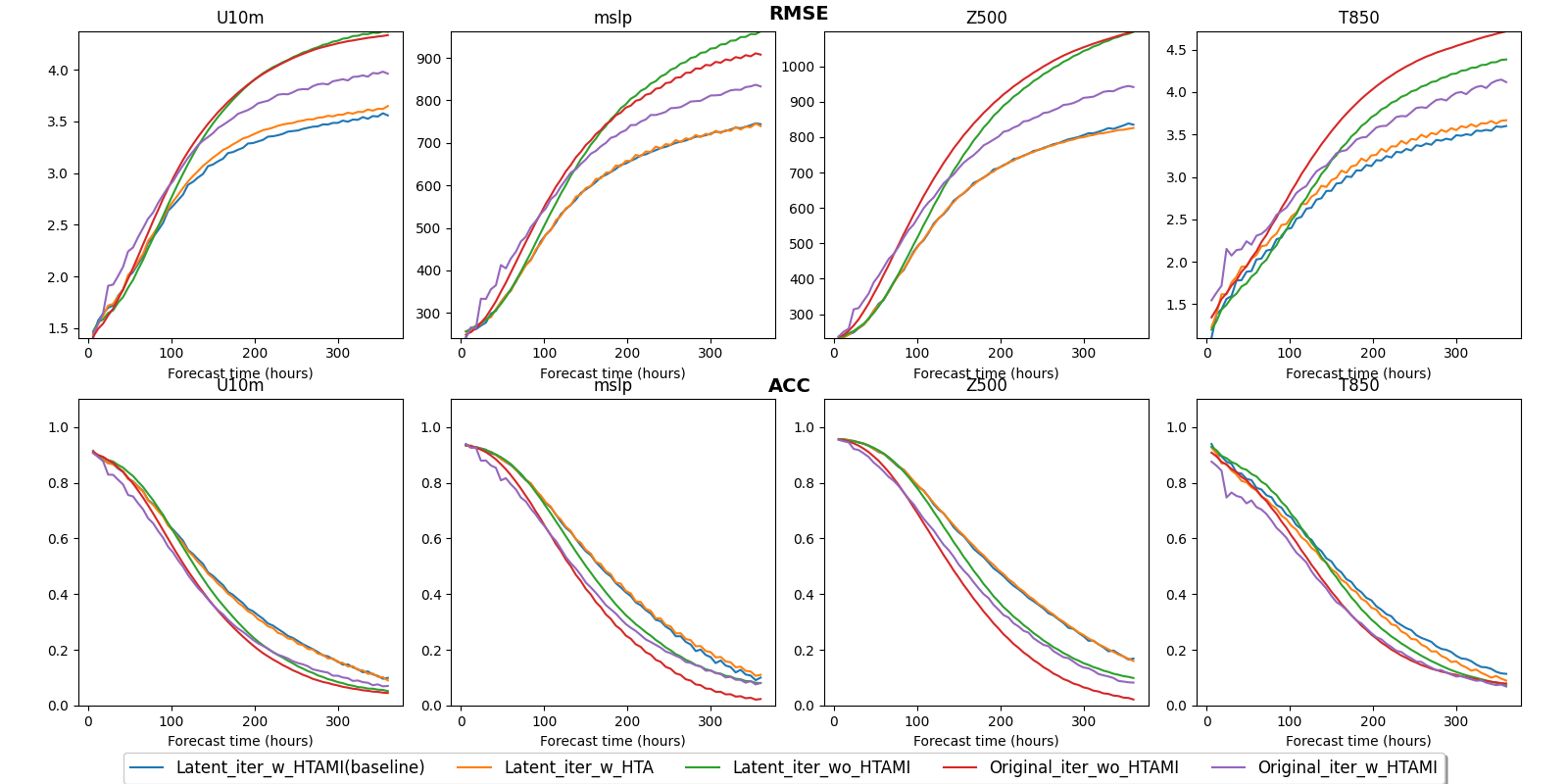}
    \caption{The ablation results of iteration manner on the global 5.625° resolution dataset. The displayed variables include U10m, mslp, Z500 and T850. ‘baseline’ means the proposed network in Section \ref{sec_method}. ‘Latent\_iter\_w\_HTAMI’ means perform the iterative prediction in the latent space with the HTAMI algorithm. ‘Original\_iter\_wo\_HTAMI’ means perform the iterative prediction in the original key variables space without the HTAMI algorithm. And ‘HTA’ is the iterative algorithm in \cite{bi2023accurate}.}
    \label{fig:Original_iter_w_HTAMI}
\end{figure*}

To validate the effectiveness of latent space iterative prediction, we conduct ablation studies to compare the performance differences between latent space iteration and original variable space iteration. Notably, latent space iteration refers to the predictive network framework in Fig. \ref{fig:network}, which performs iterative predictions in the latent space. In contrast, the original variable space iteration obtains multiply lead times' key variable predictions by iterating an integrated encode-prediction-decode network. We also test the effectiveness of incorporating the HTAMI into prediction model. Without HTAMI, as shown by the red and green lines in Fig. \ref{fig:Original_iter_w_HTAMI}, the latent feature iterative method effectively improve the predictive performance of all key variables compared to the original space iteration, particularly in the short to medium term. Although the RMSE index for the variable mslp deteriorated in the long term, the ACC index for this variable that with the latent space iteration is still superior to that of the original space iteration. 

After incorporating HTAMI, as indicated by the quantitative results of the blue and purple lines in Fig. \ref{fig:Original_iter_w_HTAMI}, all prediction results of key variables iterated in the latent space still outperform those in the original variable space in terms of both RMSE and ACC indices. In addition, as the lead time increase, this performance gap widen, suggesting that the LPM accurately capture the evolutionary pattern of latent features that is correlated with the key variables.

Furthermore, by observing the prediction performance after incorporating the HTAMI algorithm, as shown by the red/purple and blue/green lines in Fig. \ref{fig:Original_iter_w_HTAMI}, it can be seen that the long-term forecast performance significantly improves while short-term prediction performance slightly decreases. We also compared our HTAMI algorithm with the HTA algorithm in \cite{bi2023accurate}, as shown by blue and orange lines in Fig. \ref{fig:Original_iter_w_HTAMI}. And it could be inferred that our algorithm effectively increase the predictive performance at different lead times, with particularly significant improvements in the U10m and T850 prediction results. The above ablation studies fully validate the effectiveness of the latent space iteration strategy and the designed HTAMI algorithm.

\subsubsection{The effect of the constructed loss functions}

\begin{figure*}[t]
\centering
    \includegraphics[width=1\textwidth,trim=0 0 0 0,clip]{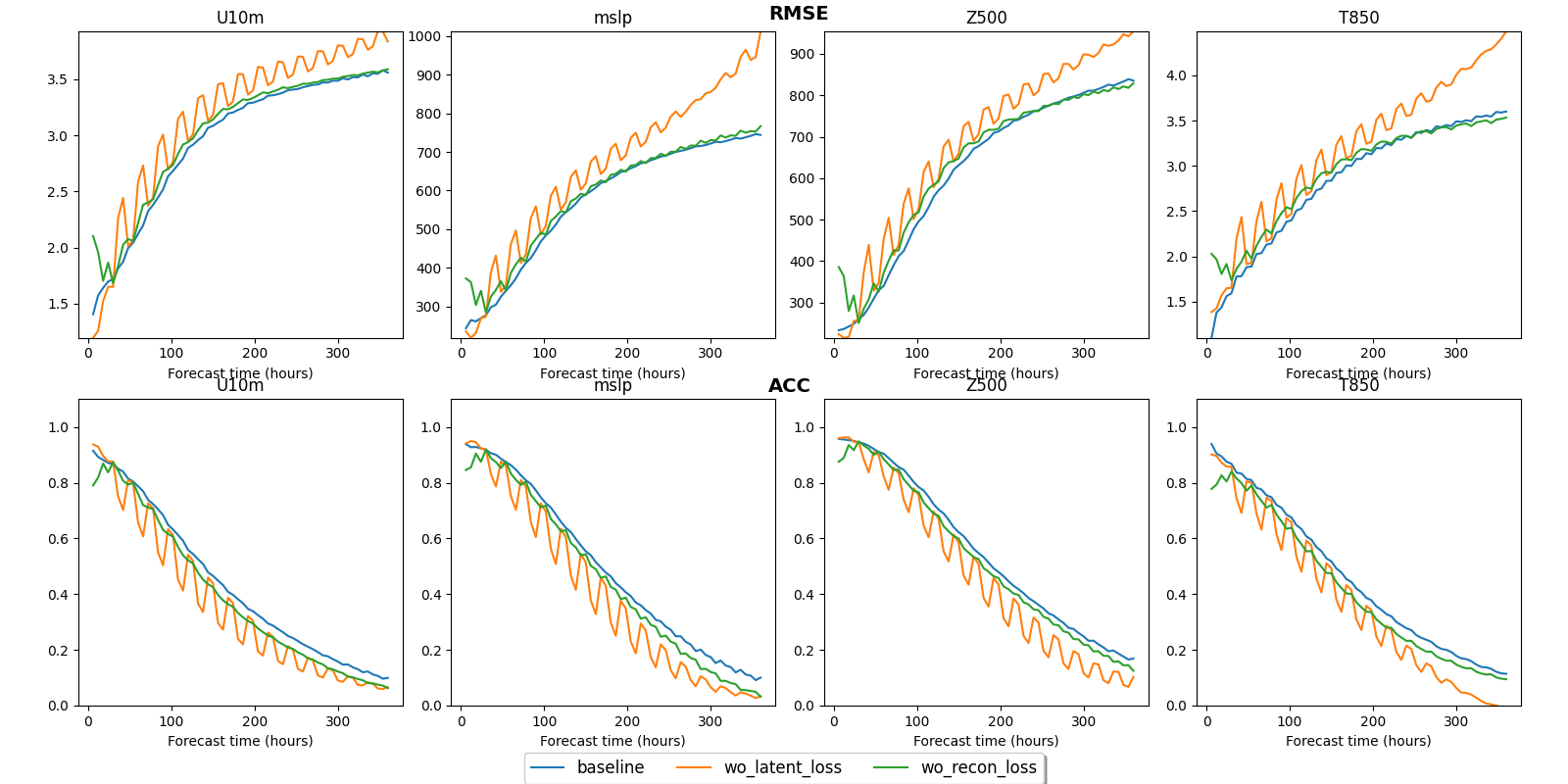}
    \caption{The ablation results of loss functions on the global 5.625° resolution dataset. The displayed variables include U10m, mslp, Z500 and T850. ‘wo\_latent\_loss’ means without the latent loss function in Eq. \ref{MSE_latent}. ‘wo\_recon\_loss’ means without the reconstruction loss function in Eq. \ref{MSE_recons}. ‘baseline’ means that encompass all loss functions.}
    \label{fig:wo_loss}
\end{figure*}

To verify the effectiveness of the proposed latent feature consistency and reconstruction loss functions, that are Eq. \ref{MSE_latent} and Eq. \ref{MSE_recons}, we conduct ablation studies. As shown in Fig. \ref{fig:wo_loss}, it can be inferred that both proposed loss functions significantly enhance the prediction performance across all lead time ranges. The latent feature consistency had a notable improvement on temporal correlation and long-term performance. Without this loss function, the RMSE and ACC indices exhibit significant fluctuations at adjacent lead times. And the reconstruction loss primarily improve the predictive performance in the short to medium term according to its RMSE index. Therefore, both constructed loss functions all effectively improve the prediction performance.

\subsection{Visualize the importance of more input variables}

\begin{figure*}[t]
\centering
    \includegraphics[width=1\textwidth,trim=100 50 100 75,clip]{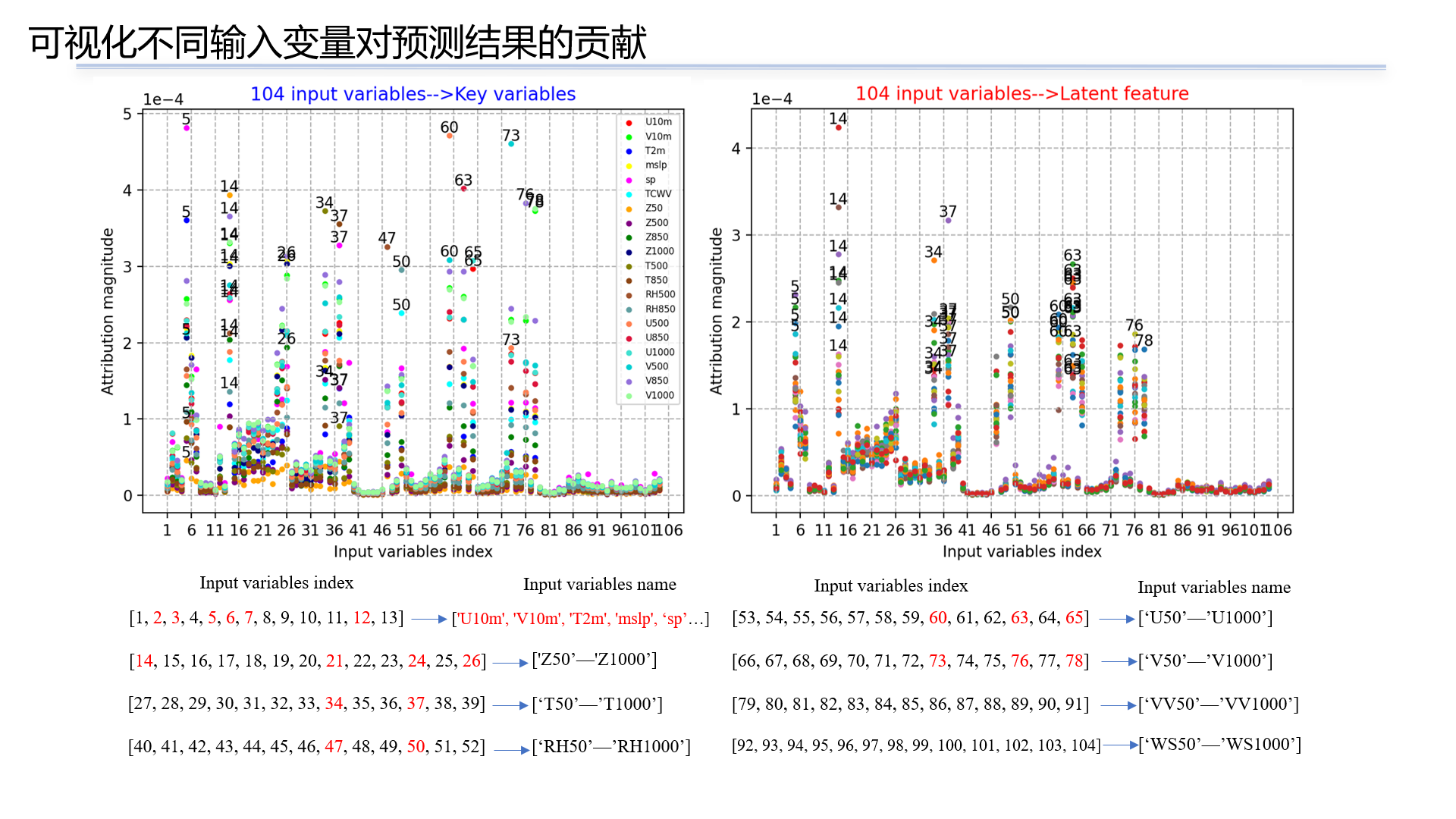}
    \caption{Attribution of input variables to the predicted key variables and encoded latent features. ‘104 input variables$\rightarrow$Key variables’ means attribution of 104 input variables to the predicted Key variables in lead time step 1. We also visualize the contribution of 104 input variables to latent feature. For each prediction variable, we label the top two feature indices with the largest attribution magnitude among the input variables.
    }
    \label{fig:grad}
\end{figure*}

Fig. \ref{fig:Original_iter_w_HTAMI} demonstrate that input more variables and perform the iterative prediction in latent space could improve the prediction performance greatly. In order to visualize the contribution of multiple input variables, we utilize the 'Integrated Gradients' method \cite{pmlr-v70-sundararajan17a} to attribute predictions to the input variables, as shown in left part of Fig. \ref{fig:grad}. 

It can be inferred that each key variable has the greatest contribution magnitude to itself. For example, The variable V500 has the greatest predictive contribution to itself with a attribution magnitude of 4.6e-4. And V1000 has the greatest predictive contribution to itself, as well as to the V10m variable, reaching a magnitude of 3.8e-4 (these two variables are very similar).
In addition, some input variables, such as z50-z1000 (with index 14-26) and T50-T1000 (with index 27-39), also exhibit pretty high attribution magnitudes. These variables with high attribution magnitudes are strongly correlated to the key variables, as observed in Fig. \ref{fig:corr_mat} of the introduction section. As shown in the right part of Fig. \ref{fig:grad}, it could also be seen that key variables and their correlated variables have the high attribution magnitudes, which means these variables other than uncorrelated variables are encoded into the latent feature. These results of attribution magnitudes indicates that not only key variables, but their related variables feature make contribution to the prediction of the key variables, which confirms the correctness of our motivation and effectiveness of the constructed prediction network.

\subsection{Complexity analyse}

\begin{table}[t]
\centering
\caption{Average computational complexity on ERA5 dataset}\label{tab1}%
\begin{tabular}{cccc}
\hline
model & \makecell[c]{Parameters \\ (M)} & \makecell[c]{Training time \\(h/2gpus)} & \makecell[c]{Inference time \\ (s/(60time steps×1gpu))}\\
\hline
SimVP & 12.530 & 4.896 & 5.304 \\ 
SimVP\_HTA & 2.960 & 2.467 & 1.206 \\
FourcastNet & 39.961 & 4.287 & 1.131  \\
FourcastNet\_HTA & 114.693  & 3.475 & 1.350  \\
PredRNNv2  & 30.703 & 3.628 & 0.829  \\
Ours& 176.984 & 6.487 & 3.138   \\
\hline
\end{tabular}
\end{table}

To validate the efficiency of the proposed method, we compare the computational complexity of different methods, including the number of network parameter, training and inference time. The training is conducted on two Nvidia RTX 3090 GPUs, while the inference was performed on a single 3090 GPU. Despite our method having a larger number of parameters and costing more training time compared to other methods, the inference time is shorter than the SimVP method by 2.166s. Additionally, above comparative experiments demonstrated that our method achieved significantly better prediction results compared to other methods. This indicates that our method achieves the best balance between effectiveness and efficiency.

\section{Conclusion}\label{sec13}

In this research, to efficiently and effectively capture the evolutionary patterns of the key variables, we input more variables and encode the variable features that is correlated with key variables into the low-dimensional latent features. And we perform the iterative prediction in this latent space and the resulted key variables are decoded from these predicted latent features. And we construct several loss functions to get more accurate prediction results.
We also improve the original HTA algorithm by inputting more time steps. The comparative experimental results on ERA5 dataset demonstrate the superiority of the proposed prediction framework in lead times up to 15 days. And the ablation studies verify the effectiveness of each component in the proposed framework. Nevertheless, as the iteration of LPM, the predicted key variables suffer from the blurring effect, which is the research direction in the future.






\bmhead{Acknowledgments}

We express our gratitude to Weatherbench2 \cite{rasp2024weatherbench} for providing the processed ERA5 dataset and related codes, as well as to OpenSTL \cite{NEURIPS2023_dcbff44d} for offering the codes of multiple predictive methods.

\section*{Declarations}

\begin{itemize}
\item Funding
\item Conflict of interest/Competing interests

The authors declare no competing interests.

\item Ethics approval 

Not applicable
\item Consent to participate

Not applicable
\item Consent for publication

Not applicable
\item Data availability

The datasets utilized in this study are all accessible on the Weatherbench2 website \href{https://console.cloud.google.com/storage/browser/weatherbench2}{https://console.cloud.google.com/storage/browser/weatherbench2}.
\item Code availability 

The codes could be available at \href{https://github.com/rs-lsl/Kvp-lsi}{https://github.com/rs-lsl/Kvp-lsi}
\item Authors' contributions
\end{itemize}







\begin{appendices}

\section{}\label{secA1}

\subsection{Overall framework}

To effectively learn complex temporal evolutionary patterns of key variables with weak correlations, we input more atmospheric variables to extract latent feature correlated to the key variables and to accurately capture their evolutionary patterns in this latent space. 

Therefore, we construct an encode-predict-decode framework, as shown in Fig. \ref{fig:network}. Initially, to derive features associated with the key variables, we develop an encoder that adaptively extracts atmospheric latent features strongly correlated with the key variables from the inputs of multiple atmospheric variables:
\begin{align}
h_{-1:0} = \varepsilon (x_{-1:0})
\label{encoding}
\end{align}
\noindent where $x_{-1:0}$ means the original atmospheric variables in time steps -1:0, and $\varepsilon$ represents the encoder. $h_{-1:0}$ is the encoded latent feature in time steps -1:0.

Subsequently, the predictor iteratively forecasts these latent features, guided by the original atmospheric variables at the corresponding lead time steps:
\begin{align}
\hat{h}_{i} = LPM (h_{i-2:i-1}), \quad  i=1,2...n
\label{pred}
\end{align}
\noindent where $\hat{h_{i}}$ is the predicted latent feature in lead time step $i$, and the $LPM$ means the latent feature prediction model that take the previous two time steps’ latent feature $h_{i-2:i-1}$ as input. Note that the number $n$ differs between the training and inference stages. The training stage involves fewer iterations, while the inference stage involves much more iterations.

At last, the iterative predicted atmospheric latent features are then fed into a decoder to yield the forecasting values of the key variables at the respective lead time:
\begin{align}
\hat{k}_{1:n} = De (\hat{h}_{1:n})
\label{decoder}
\end{align}
\noindent where $\hat{k}_{1:n}$ is the resulted key variables and $De()$ means the decoder. $\hat{h}_{1:n}$ is derive from Eq. \ref{pred}.


\subsubsection{Encoder}


To extract latent features strongly correlated with key variables, we construct an encoder based on channel-wise attention mechanism. Initially, we calculate the correlation coefficient matrix $ch\_corr_{C\times c}$ in the channel dimension for the input variables and the key variables among them. Then, through absolute value and summation operations, we obtain the correlation score coefficients $ch\_corr_{C\times 1}$ between all input bands and the key variables. 
\begin{align}
ch\_corr_{C\times c} = x_{C\times hw} \cdot k _{hw\times c }
\label{ch_multiply}
\end{align}
\begin{align}
ch\_corr_{C\times 1} = Mean(Abs(ch\_corr_{C\times c}))
\label{ch_mean}
\end{align}
\noindent where $x_{C\times hw}$ is the original input variables and $C$ is its channel number. $h$ and $w$ represent the number of grid cells of the input variables along the longitudinal and latitudinal directions, respectively. $ch\_corr_{C\times c}$ is the computed correlation coefficient matrix in the channel dimension through matrix multiplication. $Abs()$ is the operation of taking the absolute value and $Mean()$ is to compute the mean value.

A higher coefficient in $ch\_corr_{C\times 1}$ indicates that the corresponding channel in $x_{C\times hw}$ has a stronger overall correlation with the $k_{hw\times c }$ and should be given more attention. Subsequently, we multiply the $x_{C\times hw}$ by $ch\_corr_{C\times 1}$ to perform channel weighting after linearly transforming this correlation coefficient, thus highlighting atmospheric features correlated with the key variables $k_{hw\times c }$.

After channel weighting, the input features are passed through four consecutive multi-scale residual blocks to obtain the latent features.
\begin{align}
h_{d\times hw} = MSR\_block^4(\hat{x}_{C\times hw})
\label{MSR}
\end{align}
\noindent where $h_{d\times hw}$ is the encoded latent feature with channel number of $d$ ($d\ll C$). $MSR\_block^4()$ represents four multi-scale residual blocks and $\hat{x}_{C\times hw}$ is the channel-weighted input variables. Note that $d$ is set to 24 in this research.

As shown in Fig. \ref{fig:encoder}, each multi-scale residual block consists of two convolutional modules with different local feature receptive fields, where the kernel sizes are 3 and 5, respectively. Designing convolutional modules with two different kernel sizes allows for the full extraction of multi-scale spatial detail features. The results of these two convolutional blocks are then concatenated in channel dimension, followed by a 1×1 convolution layer before output. Each convolutional module includes convolution + GroupNorm + SiLU activation function to enhance the network's ability to learn non-linear feature transformations. Finally, a residual connection is added to accelerate convergence and mitigate the gradient vanishing/exploding problem \cite{He_2016_CVPR}.

\subsubsection{Prediction model}


After obtaining the encoded latent features $ h_{-1:0} $ (this subscript denotes
two time steps), we input them into the 3-dimensional convolution and vision transformer (3d-ViT) based prediction model to predict the next time step's latent feature $\hat{h}_{1}$. Note that we input 2 time steps into the LPM to enhance the temporal relation between the input and output. Additionally, we include the corresponding time embedding to enhance sensitivity to different time steps and add the constant embedding. These three components are concatenated along the channel dimension to serve as the model input. 
\begin{align}
h_{(2d+dc+dt)\times hw} = Concat(h_{2d\times hw}, x\_time_{dt\times hw}, x\_const_{dc\times hw})
\label{Concat}
\end{align}
\noindent where $h_{2d\times hw}$ is the input latent feature and $2d$ means the two time steps × channel number of latent feature. $x\_time_{dt\times hw}$ is the time embedding features after aligning the spatial dimension with $h_{2d\times hw}$. $x\_const_{dc\times hw}$ is the constant embedding features. $dt$ and $dc$ are their channel number. $Concat$ means the concatenation operation along the channel dimension. 

\begin{figure*}[t]
\centering
    \includegraphics[width=0.8\textwidth,trim=0 200 500 0,clip]{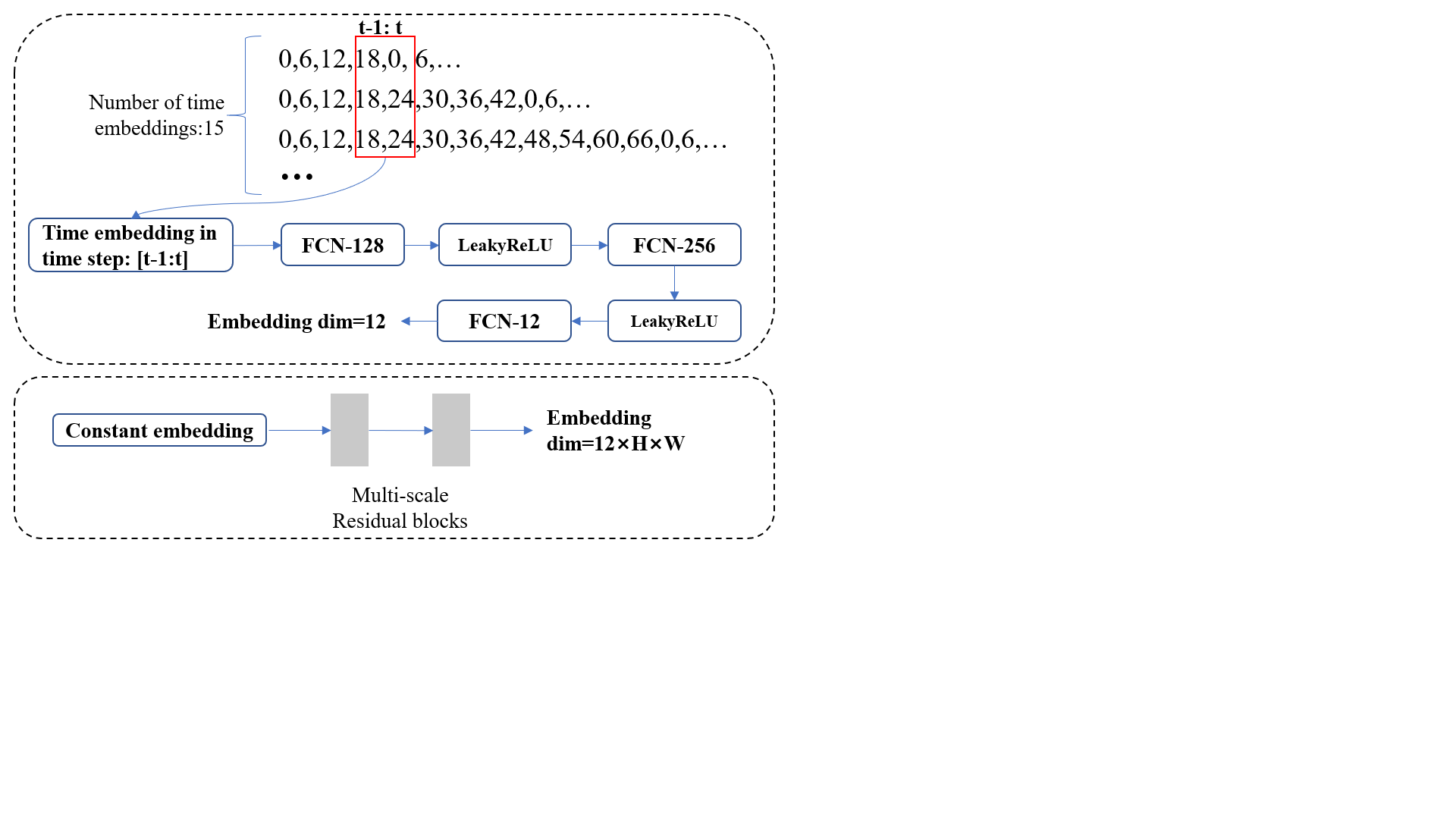}
    \caption{The overall prediction model structure. The blue box indicates the forward propagation process ‘encode-predict-decode’. The green box contains the latent feature loss function component, and the bottom part represents the key variable loss function. 
    }
    \label{fig:embedding}
\end{figure*}
Due to the temporal resolution of the data being 6 hours and in order to forecast key variables for the next 15 days, we constructed the multi-cycles time embedding data, as shown in top of Fig. \ref{fig:embedding}. These embedding data have different repeating cycles, totaling 15 cycles ranging from 1 day to 15 days. This ensures that the prediction model is sensitive to atmospheric circulation activity with different cycles. The time embedding processing module, as illustrated in Fig. \ref{fig:embedding}, takes two time embeddings as input and through cascaded fully connected layers and LeakyRuLU activation function, and the output is with a dimension of 12. The "FCN-128" refers to a fully connected network with an output dimension of 128. For constant feature, they are pass through two multi-scale residual modules and output embedding features with 12 channels. The multi-scale residual module is depicted in Fig. \ref{fig:encoder}.

The input features $h_{(2d+dc+dt)\times hw}$ first pass through four multi-scale residual blocks (as in Fig. \ref{fig:encoder}) to aggregate these feature information, as shown in the left part of Fig. \ref{fig:LPM_network}, allowing temporal and constant information to be fully embedded into the latent features. We then employ a 3-dimensional convolution (3Dconv) module to mine the temporal correlation between two time steps’ latent features, aiming to enhance the temporal correlation with the predicted latent feature:
\begin{align}
h_{2d\times hw} = Res\_{block}^4(h_{(2d+dc+dt)\times hw})
\label{Resblock}
\end{align}
\begin{align}
h_{2d\times hw} = 3D\_conv(h_{2d\times hw})
\label{3dconv}
\end{align}
\noindent where $Res\_{block}^4()$ represents four residual blocks and $3D\_conv()$ is the 3Dconv module. 3Dconv module include two 3D convolutional layers and a SiLU activation function between them.

\begin{table}[t]
\centering
\caption{Major hyperparameter of the ViT}\label{tab_hyper}%
\begin{tabular}{cc}
\hline
Hyperparameter & Value \\
\hline
patch size  & [4,4]   \\
embedding dimension & 768  \\
depth & 8   \\
number of heads & 12  \\
mlp ratio & 4   \\
\hline
\end{tabular}
\end{table}

The latent features $h_{2d\times hw}$ are then fed into ViT for feature enhancement. ViT is a variant of the transformer that can learn the global correlation of image features and enhance them. And it incorporate patch-embedding to learn the patch-wise non-local similarities, which can effectively and efficiently learn the local spatial feature and their global similarity to enhance the latent features. 

As depicted in Fig. \ref{fig:LPM_network}, to fully utilize the local feature of input feature, we use the linear projection, namely stride convolutional layers, for patch extraction and flatten the patch size dimension with the channel and time dimensions. We also add the position embedding to learn the relative position relationships between different location features:
\begin{align}
h_{emb\_dim \times \frac{h}{ps}\times \frac{w}{ps}} = patchify(h_{2d\times hw})
\label{patchify}
\end{align}
\noindent where $patchify()$ means take the stride convolution operation with a stride equal to patch size. And $emb\_dim$ is the patch-embedded feature dimension. In this research the patch size is set to 4 and feature embedding dimension-- $emb\_dim$ is 768. Actually, the $emb\_dim$ incorporate two input time steps, latent feature in different channel and local patch feature these three dimensions.

Then, we input the patch-embedded features $h_{emb\_dim \times \frac{h}{ps}\times \frac{w}{ps}}$ into L cascaded ViT blocks for feature enhancement. Each block comprises two residual modules; the first includes Layernorm and a Multi-head Attention (MHA) module. 
\begin{align}
h_{emb\_dim\times h’w’} = MHA(LN(h_{emb\_dim\times h’w’})) + h_{emb\_dim\times h’w’}
\label{MHA}
\end{align}
\noindent where $h’w’$ is equals to $\frac{h}{ps}\times \frac{w}{ps}$. $MHA$ represents the multi-head attention operation and $LN$ is the LayerNorm layer. LayerNorm could assists the model in handling inputs feature of varying scales and stabilizes gradient variations. MHA calculates globally patch-wise self-attention, effectively and efficiently enhance the input image feature. Moreover, the addition of residual connections can effectively mitigate gradient vanishing and exploding problems. Subsequently, a LayerNorm+MLP (multi-layer perceptron) residual module further enhances these features. Unlike MHA, MLP equally processes different features initially, which could focus on parts that previously received less attention:
\begin{align}
h_{emb\_dim\times h’w’} = MLP(LN(h_{emb\_dim\times h’w’})) + h_{emb\_dim\times h’w’}
\label{MLP}
\end{align}
\noindent where $MLP$ represents the multi-layer perceptron network. Finally, the enhanced latent features are restored to the same shape as the input through patch recovery, and after passing through four multi-scale residual blocks, the predicted next time step’s latent features are output.
\begin{align}
h_{2d\times hw} = Patch\_recovery(h_{emb\_dim\times h’w’})
\label{Patch_recovery}
\end{align}
\begin{align}
\hat{h}_{d\times hw} = Res\_{block}^4(h_{2d\times hw})
\label{MLP}
\end{align}
\noindent where $Patch\_recovery()$ is the patch recovery operation and it include a LayerNorm, a full connected layer and reshape operation.

Upon obtaining the predicted latent features, the LPM takes the latent features of time step t and the predicted latent features of lead time step t+1 as input and forecast the latent features of lead time step t+2, and iterates multiple rounds to obtain multiple lead times' latent feature predictions.

\subsubsection{Model iterative manner}

As the number of LPM iterations increases, cumulative errors would become increasingly severe. \cite{bi2023accurate} proposed the HTA algorithm to minimize the number of prediction model iterations, that is, by training multiple models with different time intervals. However, for this algorithm, each predictive model only takes the previous single time step as input to predict the next time step, which limits the amount of historical information that the model can receive, making it difficult for the model to capture the complex temporal dependencies between the input and output sequences. 

Moreover, the HTA algorithm in \cite{bi2023accurate} cannot directly input two time steps to predict the next one because if each prediction model is fed two consecutive time steps, it would be impractical to select an appropriate prediction model during the inference stage. For instance, taking two time steps with a 6-hour interval as input of a 24-hour interval prediction model would not be suitable. Similarly, if one LPM initially takes two time steps with a 24-hour interval as input, it would also be inappropriate to subsequently feed them into a prediction model with 6-hour interval.

Therefore, we propose the Hierarchical Temporal Aggregation with More Inputs (HTAMI) algorithm, which increases the temporal correlation of the prediction results by taking more (two) time steps as input. During the training stage, we train multiple LPMs with different time intervals, that is, taking two time steps as input with intervals of 6, 12, and 24 hours (equal to 1, 2, and 4 time steps and the corresponded LPM are LPM1, LPM2 and LPM4 in Fig. \ref{fig:iteration_manner}, respectively). During the inference stage, as depicted in Fig. \ref{fig:iteration_manner}, we first input the atmospheric variables from the previous five time steps. Then, for each lead time step, we select the optimal time interval prediction model. Algorithm \ref{iteration_algo1} details the iterative approach during the inference stage.

\subsection{Experiment}
\subsubsection{Dataset}

The validity of the proposed encoding-prediction-decoding iterative prediction framework was verified through experiments conducted on the ERA5 dataset. The ERA5 is a reanalysis dataset obtained from the European Centre for Medium-Range Weather Forecasts (ECMWF), encompassing various ground surface and atmospheric variables across multiple vertical pressure levels, spanning from 1959 to the present, with a temporal resolution of one hour and a spatial resolution of 0.25°.

Weatherbench2 \cite{rasp2024weatherbench} is an integrated and processed version of the official ERA5 data, enabling rapid downloads, facilitating comparisons between predicted results of different methods. Due to device memory constraints, we selected the processed ERA5 dataset from Weatherbench2 with a spatial resolution of 5.625° for our experiments. This dataset's temporal resolution was downsampled to 6 hour intervals, and the key variables for prediction are listed in Table \ref{tab_key_var}, with a total forecasting period of 15 days (60 lead time steps).

Our proposed model is designed to benefit to more atmospheric variables than key variables, so we input seven variables at thirteen vertical pressure levels, including 'geopotential', 'temperature', 'specific humidity', 'u component of wind', 'v component of wind', 'vertical velocity', 'wind speed'. The thirteen pressure levels encompass 50, 100, 150, 200, 250, 300, 400, 500, 600, 700, 850, 925 and 1000 hPa. Additionally, thirteen surface variables are included. Except that in key variables, the surface variables also include ‘total precipitation 6hr’, ‘10m wind speed’, ‘toa incident solar radiation’, ‘toa incident solar radiation 12hr’, ‘toa incident solar radiation 6hr’, ‘total cloud cover’, ‘total precipitation 12hr’. So the input includes 104 temporal variables, with the key variable being among them. Constant variables, such as land-sea masks and land cover types, are also incorporated. To enhance the model's sensitivity to different input times, we input multiple time embeddings with varying cyclical periods. As the data ranges of different variables varied, each variable was normalized before entering the model – that is, mean-subtracted and divided by the standard deviation. The outputs for the model's key variables were then reversely normalized.

The datasets for model training, validation, and testing were segmented into the periods of 2000-2017, 2018-2019, and 2020, respectively. The data from 2020 was only used for testing at the end, and only the validation set was used during the model construction and validation phase.

\subsubsection{Metric indices}

This study employ two performance metric indices to measure the consistency between the predictive key variables and their true values in ERA5, including Root Mean Square Error (RMSE) and anomaly correlation coefficient (ACC). RMSE evaluates the overall contour consistency between the predicted results and the true values. In contrast, ACC employs historical climate averages to compute a similarity index of the local detailed feature between the predicted results and the true values. Furthermore, both indices incorporate a weighted computation across different latitudes. A lower RMSE index and higher ACC index indicate more accurate prediction results. Note that we utilize the 2020 year test set to conduct comparisons of different methods, and all methods receive the five time steps before 00:00 and 12:00 UTC daily and predict the key atmospheric variables for 360 hours (60 lead time steps).

\subsubsection{Details of reproducing other methods}\label{secA11}

Due to the model limitations of the compared prediction methods, these methods' input features could only include the key variables to be predicted. The total input time steps of PredRNNV2, SimVP and FourCastNet are all set to 5 in training and inference stages, which is same as our model that takes five time steps as input during the inference phase (even our model takes two time steps as input in the training stage). This ensures the fairness that different methods can obtain input features with the same time steps. The time step of each output from these models are all set to 1. For the SimVP\_HTA and FourCastNet\_HTA methods, due to the limitation of HTA algorithm, their input time steps are set to 1.

During the training phase, the increment schedule of all lead time steps of these methods are same as ours--[2,4,6,8]. And to make this increment schedule compatible with our methods to ensure the fairness, the epoch of these methods are all set to 65. In this case, the corresponding epochs schedule that with different lead time steps could also following our method--[0-50, 50-55, 55-60, 60-65]. Their learning rates and decay schedules are following the original settings.

\subsubsection{Comparative results on other key variables}

\begin{figure*}[h]
\centering
    \includegraphics[width=1\textwidth,trim=0 0 0 0,clip]{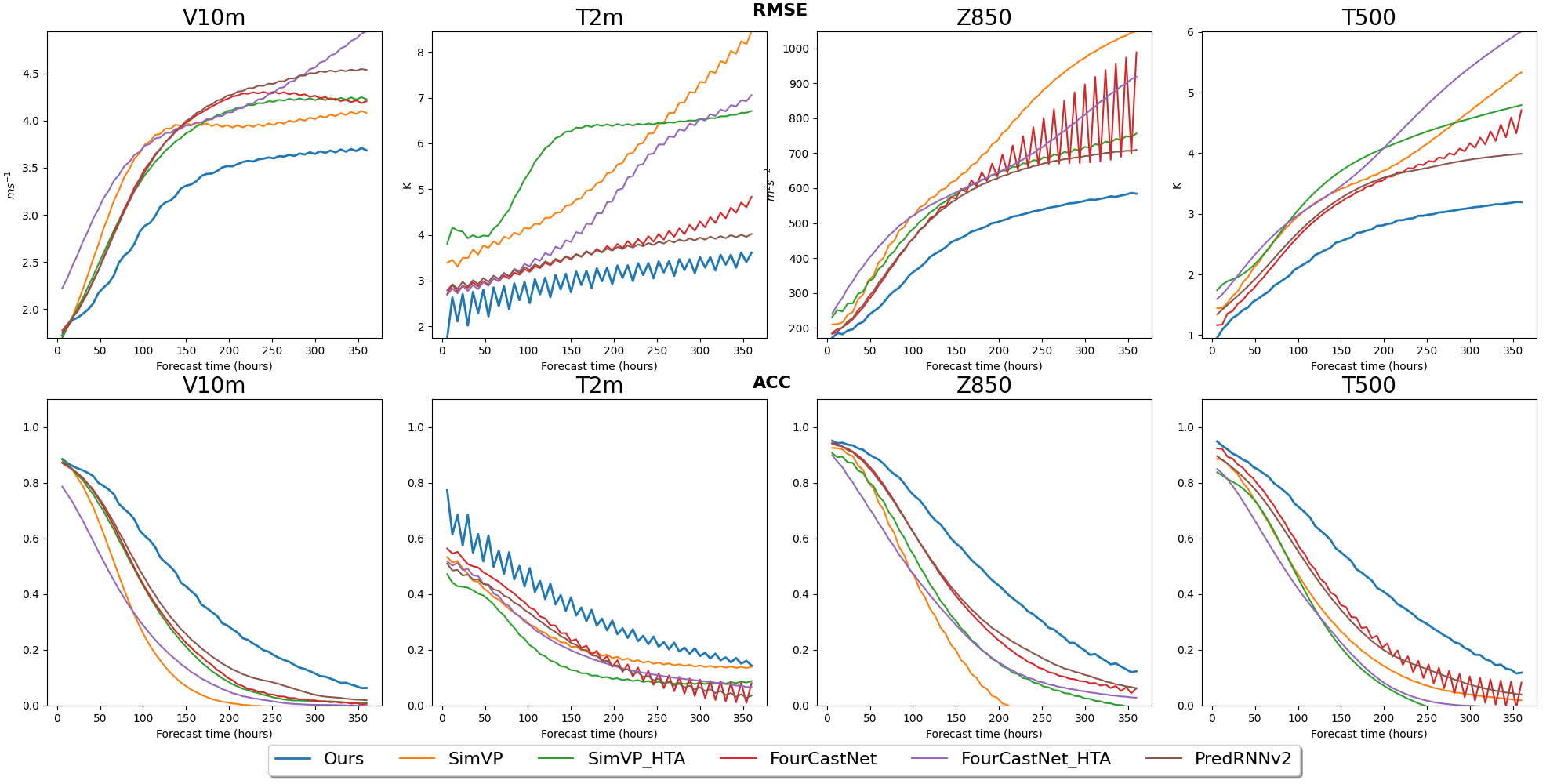}
    \caption{The averaged quantitative comparison results on the global 5.625° resolution dataset. The displayed key variables include 10m v component of wind (V10m), 2m temperature (T2m), geopotential at 850hpa (Z850) and temperature at 500hpa (T500).}
    \label{fig:metric_V10m}
\end{figure*}

To verify the superiority of the proposed prediction model, we also display other key variables, including 10m v component of wind (V10m), 2m temperature (T2m), geopotential at 850 hpa (Z850), and temperature at 500 hpa (T500). As shown in Fig. \ref{fig:metric_V10m}, compare with other methods, our method's forecasted results on these four variables were significantly better across all lead times in terms of RMSE and ACC indices. Despite the fluctuations in our method's T2m prediction results, our method still outperforms other methods. 
And the performance gap in terms of RMSE index between our approach and other methods tends to increase as the lead time increase.

\begin{figure*}[h]
\centering
    \includegraphics[width=1\textwidth,trim=0 0 0 0,clip]{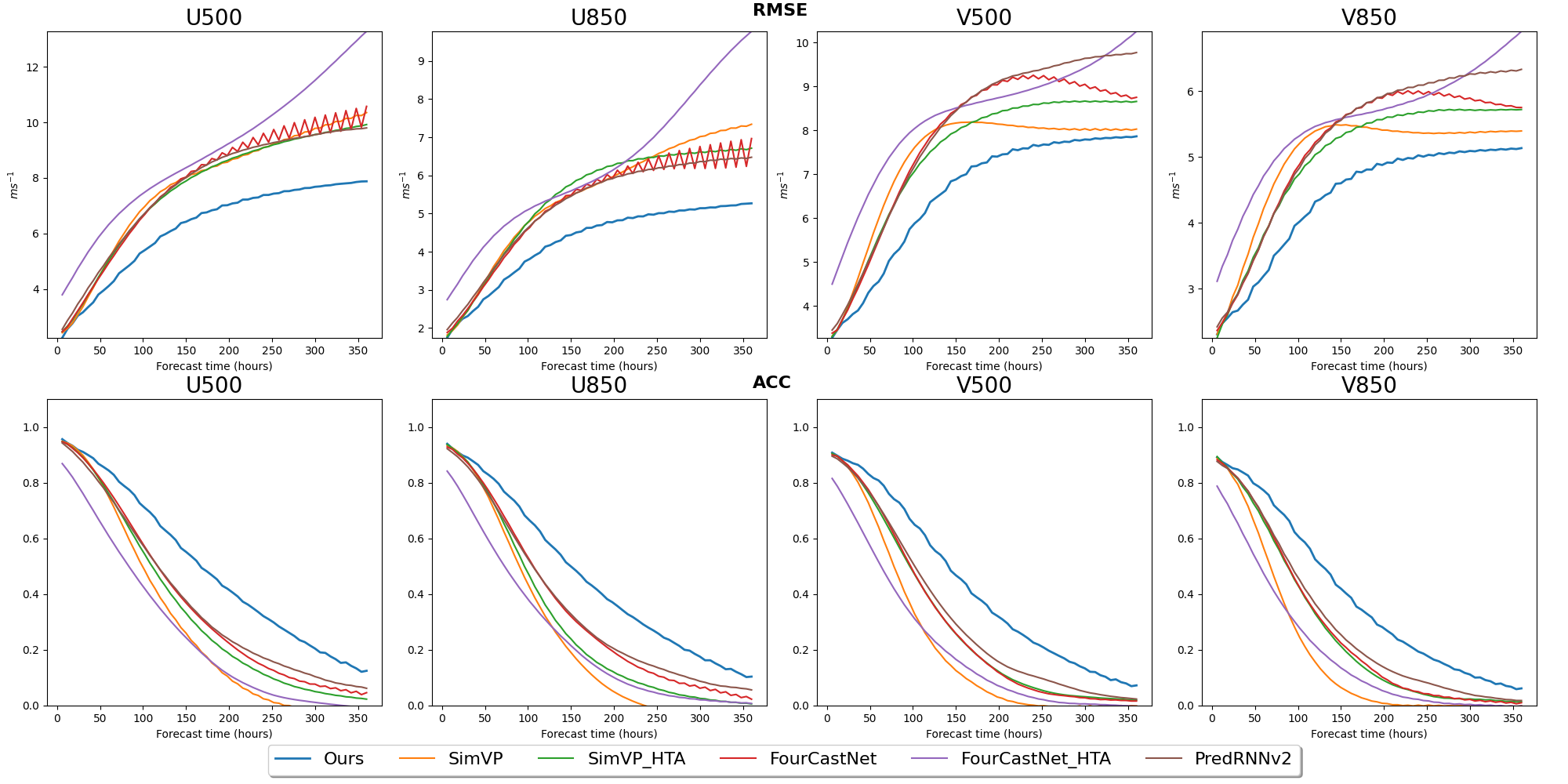}
    \caption{The averaged quantitative comparison results on the global 5.625° resolution dataset. The displayed key variables include u component of wind at 500 hpa (U500), u component of wind at 850 hpa (U850), v component of wind at 500 hpa (V500), v component of wind at 850 hpa (V850).}
    \label{fig:metric_UV}
\end{figure*}

\begin{figure*}[h]
\centering
    \includegraphics[width=0.95\textwidth,trim=40 60 0 20,clip]{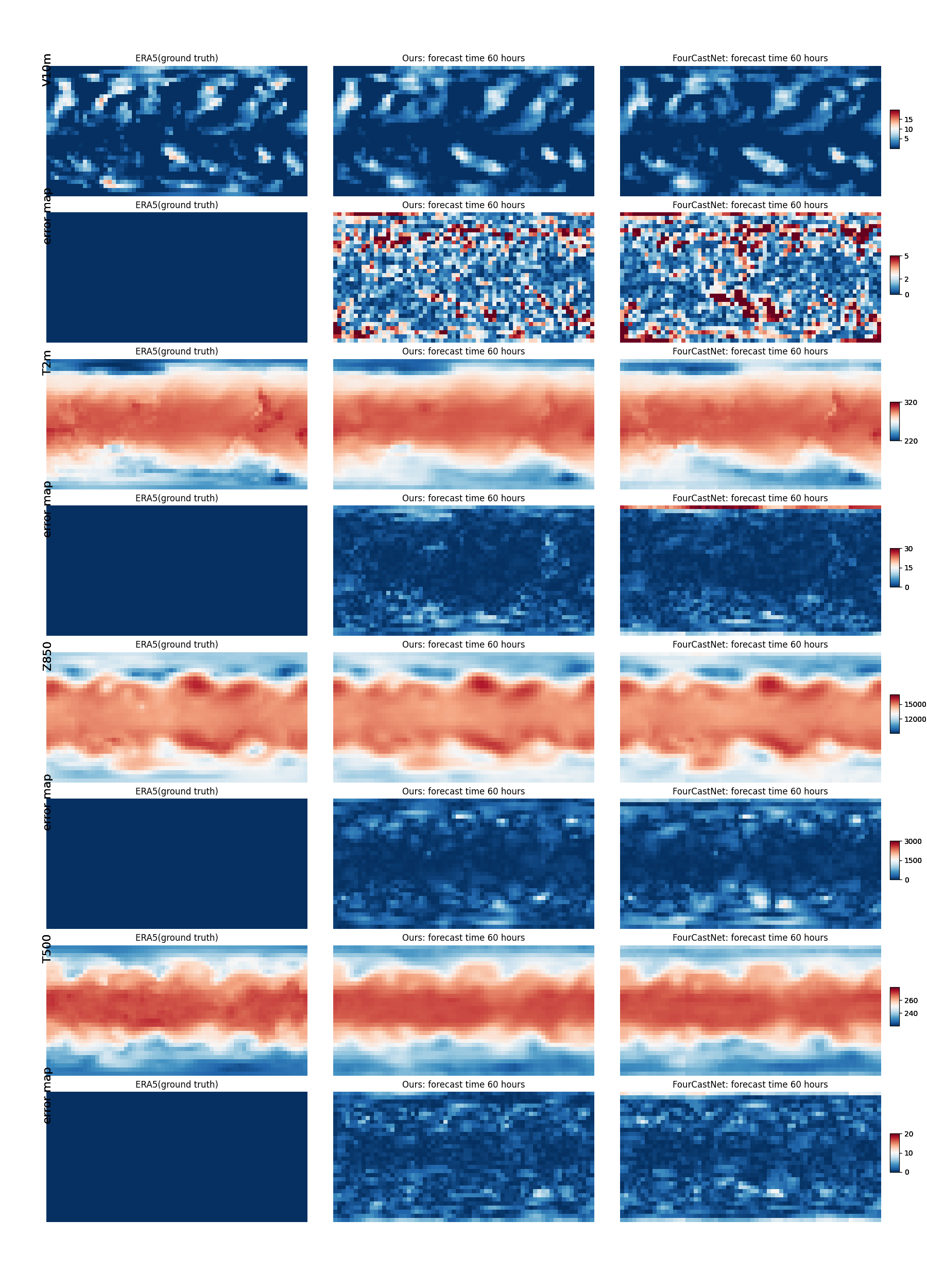}
    \caption{The predicted results of lead time 60 hours that take 5 time steps before 2020.02.25.12:00 as the model input. The displayed key variables include V10m, T2m, Z850 and T500.}
    \label{fig:quali_fig_V10m}
\end{figure*}

We also present the qualitative results for the above four key variables at lead time of 60-hour, as shown in Fig. \ref{fig:quali_fig_V10m}. 
For the V10m variable, our method more accurately predict the trend of wind belt changes in the South Pacific. For Z500, our method also more accurately predicted the changes in the mid-pressure area and boundaries in the South Pacific.
Therefore, our method achieves forecasting outcomes superior to those of FourCastNet.

\subsubsection{Comparative results on last key variables}

To further verify the advantage of our method, we also display other key variables, including u component of wind at 500 hpa (U500), u component of wind at 850 hpa (U850), v component of wind at 500 hpa (V500), v component of wind at 850 hpa (V850). Compare with other methods, as shown in Fig. \ref{fig:metric_UV}, our method's forecast results on these four variables are evidently better across all forecast durations in terms of both RMSE and ACC performance indices. 

\begin{figure*}[h]
\centering
    \includegraphics[width=0.95\textwidth,trim=40 60 0 20,clip]{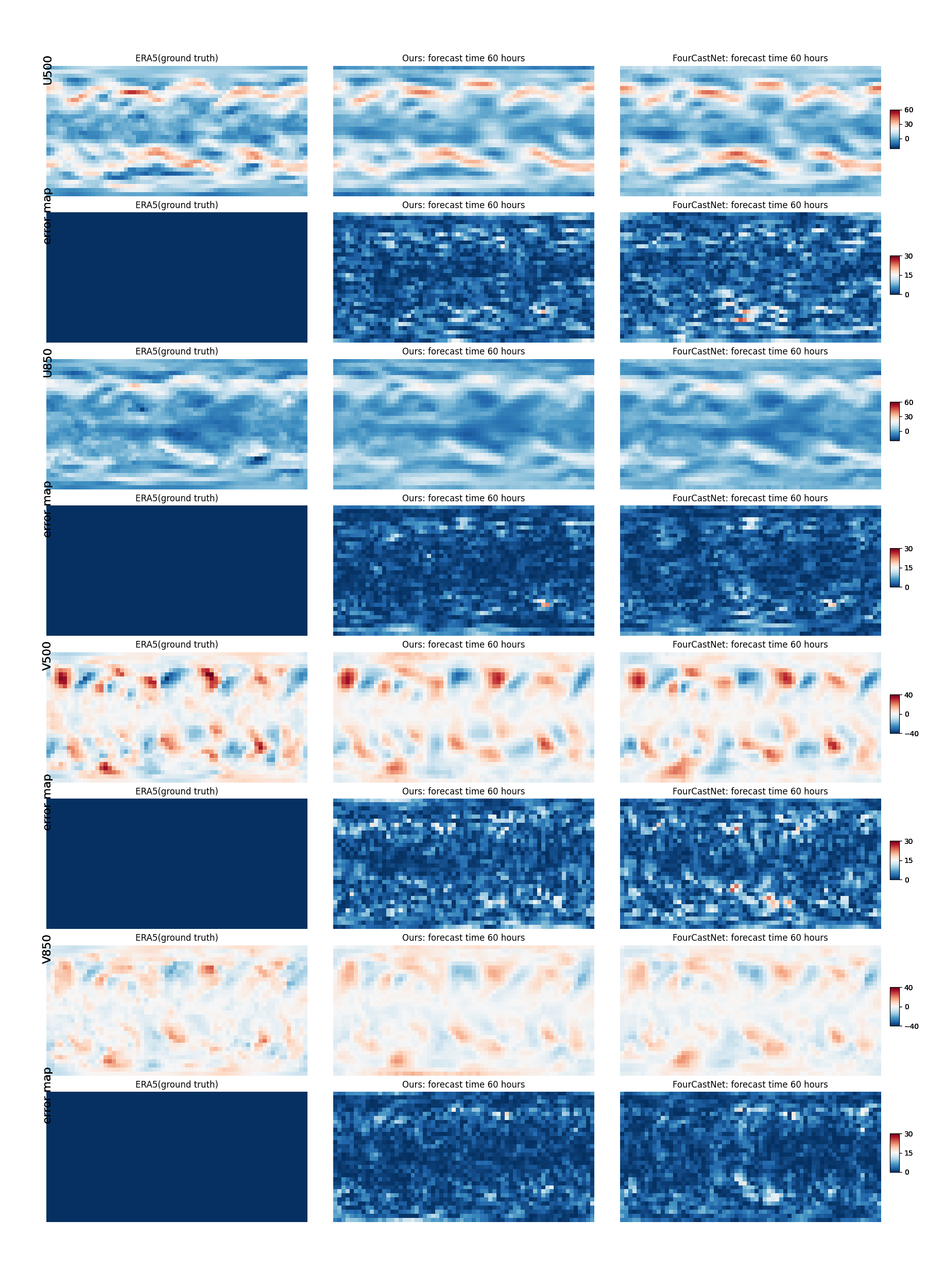}
    \caption{The predicted results of lead time 60 hours that take 5 time steps before 2020.02.25.12:00 as the model input. The displayed key variables include U500, U850, V500, V850.}
    \label{fig:quali_fig_UV}
\end{figure*}

We also presented the qualitative results for the another four key variables at lead time of 60-hour, as shown in Fig. \ref{fig:quali_fig_UV}. As shown in Fig. \ref{fig:quali_fig_UV}, it could be seen that our method achieve more accurate prediction results in south pacific area for the above four variables, which could be chearly seen from the error maps. The all above comparative prediction results fully demonstrates that the proposed latent space iterative model can effectively excavate and learn the evolutionary patterns of key variables, and also proves the superiority of our model in predicting various variables. 

\end{appendices}


\bibliography{sn-article}

\end{document}